\documentclass{article}

\usepackage{microtype}
\usepackage{graphicx}
\usepackage{subfigure}
\usepackage{booktabs} 
\usepackage{multirow}
\usepackage{hyperref}

\usepackage{algorithm}
\usepackage{algorithmic}

\usepackage[accepted]{icml2025}

\usepackage{amsmath}
\usepackage{amssymb}
\usepackage{mathtools}
\usepackage{amsthm}
\usepackage{xcolor}

\usepackage[capitalize,noabbrev]{cleveref}

\theoremstyle{plain}
\newtheorem{theorem}{Theorem}

\theoremstyle{definition}

\theoremstyle{remark}

\usepackage[textsize=tiny]{todonotes}

\usepackage{stfloats}
\newcommand\bai[1]{{\color{blue}#1}}

\icmltitlerunning{Weak-to-Strong Diffusion with Reflection}

\begin{document}

\twocolumn[
\icmltitle{Weak-to-Strong Diffusion with Reflection}

\begin{icmlauthorlist}
\icmlauthor{Lichen Bai}{h}
\icmlauthor{Masashi Sugiyama}{r,t}
\icmlauthor{Zeke Xie}{h}
\end{icmlauthorlist}

\icmlaffiliation{h}{xLeaF Lab, The Hong Kong University of Science and Technology (Guangzhou)}
\icmlaffiliation{r}{RIKEN AIP}
\icmlaffiliation{t}{The University of Tokyo}

\icmlcorrespondingauthor{Zeke Xie}{zekexie@hkust-gz.edu.cn}

\icmlkeywords{Generative Model, Diffusion Model}

\vskip 0.3in
]

\printAffiliationsAndNotice{}

\begin{abstract}
The goal of diffusion generative models is to align the learned distribution with the real data distribution through gradient score matching. However, inherent limitations in training data quality, modeling strategies, and architectural design lead to inevitable gap between generated outputs and real data.
To reduce this gap, we propose Weak-to-Strong Diffusion (\textbf{W2SD}), a novel framework that utilizes the estimated difference between existing weak and strong models (i.e., weak-to-strong difference) to bridge the gap between an ideal model and a strong model. By employing a reflective operation that alternates between denoising and inversion with weak-to-strong difference, we theoretically understand that W2SD steers latent variables along sampling trajectories toward regions of the real data distribution. W2SD is highly flexible and broadly applicable, enabling diverse improvements through the strategic selection of weak-to-strong model pairs (e.g., DreamShaper vs. SD1.5, good experts vs. bad experts in MoE). Extensive experiments demonstrate that W2SD significantly improves human preference, aesthetic quality, and prompt adherence, achieving SOTA performance across various modalities (e.g., image, video), architectures (e.g., UNet-based, DiT-based, MoE), and benchmarks. For example, Juggernaut-XL with W2SD can improve with the HPSv2 winning rate up to \textbf{90\%} over the original results. Moreover, the performance gains achieved by W2SD markedly outweigh its additional computational overhead, while the cumulative improvements from different weak-to-strong difference further solidify its practical utility and deployability. The
code is publicly available at \href{https://github.com/xie-lab-ml/Weak-to-Strong-Diffusion-with-Reflection}{github.com/xie-lab-ml/Weak-to-Strong-Diffusion-with-Reflection}.
\end{abstract}

\section{Introduction}
\label{Introduction}

In probabilistic modeling, the estimated density represents model-predicted distributions while the ground truth density reflects actual data distributions. Diffusion models~\citep{songscore}, known for its powerful generative capabilities and diversity, estimate the gradient of the log probability densities in perturbed data distributions by optimizing a score-based network, which has become a mainstream paradigm in many generative tasks~\citep{podellsdxl,guo2023animatediff}.

\begin{figure}[t]
    \centering
    \includegraphics[width=1.0\linewidth]{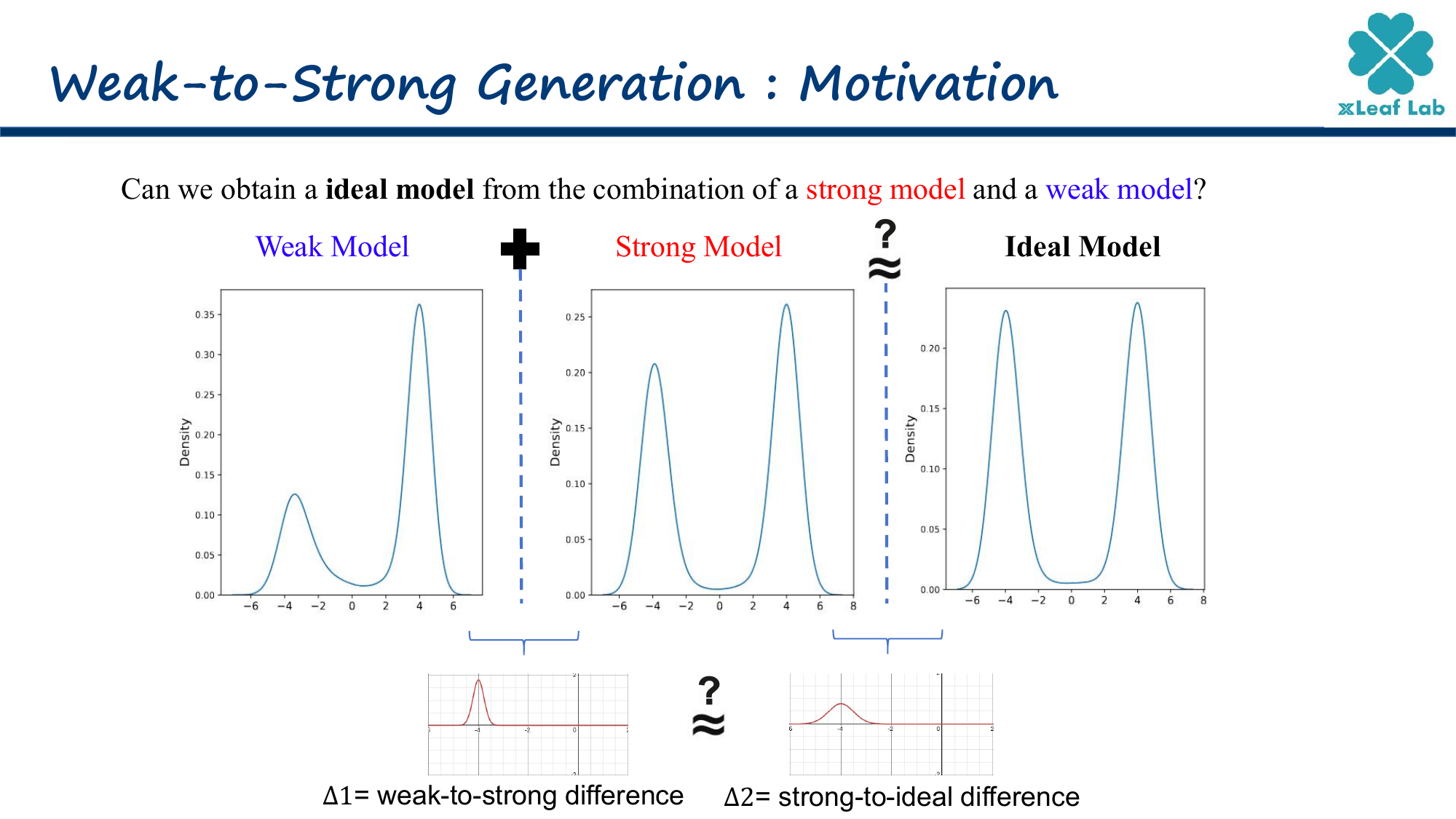}

        \vspace{-0.3cm}

    \caption{W2SD leverages the gap between weak and strong models to bridge the gap between strong and ideal models.}
    \vspace{-0.2cm}
    \label{fig:motivation}
\end{figure}

\begin{figure*}[t]
    \centering
\includegraphics[width=0.9\linewidth]{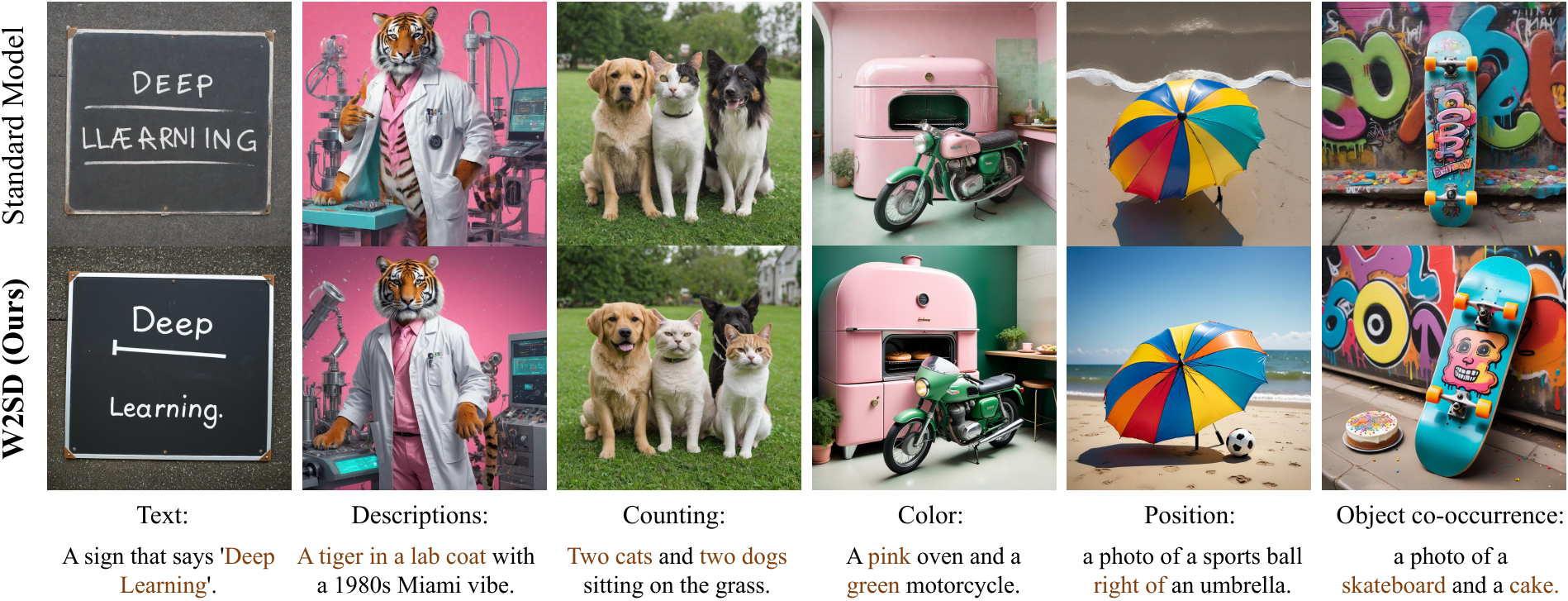}
    \vspace{-0.35cm}
    \caption{The qualitative results of W2SD demonstrate the effectiveness of our method in various aspects, such as text rendering, position, color, counting, and object co-occurrence. We present more cases in Appendix~\ref{sec:exp_res_quali}.}
    \label{fig:main_case}
\end{figure*}

To enhance the alignment between the learned and real data distributions in diffusion models, extensive research has focused on improving the gradient alignment between estimated and ground truth densities through various strategies~\citep{karras2022elucidating, song2020improved,ho2022video}. However, constrained by practical limitations including architectural design and dataset quality issues, existing diffusion models inevitably exhibit gradient estimation errors, leading to a gap between the learned and real data distributions. Depending on the magnitude of this gap, we can classify them as either ``strong'' or ``weak'' diffusion models. Specifically, due to the inaccessibility of the real data distribution, we cannot establish direct methodologies to measure this gap and effectively reduce it.

In this work, we demonstrate that the gap between the strong and weak models can be used to empirically bridge the gap between the ideal and strong models. In~\cref{fig:motivation}, we propose to use the estimated weak-to-strong difference $\Delta_{1}$ to bridge the inaccessible strong-to-ideal difference $\Delta_{2}$, thereby enhancing the strong model toward the ideal model. Further evidence supporting the viability of using $\Delta_{1}$ to bridge $\Delta_{2}$ is provided in~\cref{sec:deeper_visual_understanding}. 

We note this weak-to-strong concept has been widely applied in training processes of machine learning, with early instances such as AdaBoost~\citep{schapire2013explaining}, where the ensembling of weak models enhances the performance of strong models. Also,~\citet{burns2023weak} have demonstrated that weak models can serve as supervisory signals to assist in the alignment during the training of large LLMs. 

To empirically estimate the weak-to-strong difference, we draw upon reflective mechanisms, the process of modifying generated outputs based on prior states, which have been extensively studied in the field of LLMs~\citep{goucritic,madaan2024self,shinn2024reflexion}. Specifically, we propose Weak-to-Strong Diffusion (\textbf{W2SD}), a novel framework that leverages weak-to-strong difference to bridge the strong-to-ideal difference. By incorporating a reflective operation that alternates between denoising and inversion based on the weak-to-strong difference, we theoretically understand in~\cref{sec: theory} that W2SD guides latent variables along sampling trajectories, effectively steering them toward regions of the real data distribution. And we provide qualitative results in~\cref{fig:main_case}.

We emphasize that W2SD can be generalized to diverse application scenarios. Notably, in~\cref{sec:other_methods} we demonstrate existing inference enhancement methods including Re-Sampling~\citep{lugmayr2022repaint}, Z-Sampling~\citep{bai2024zigzag}, FreeDom~\citep{yu2023freedom} and TFG~\citep{ye2024tfg} can be reinterpreted as specialized instances of W2SD. Users can define ``weak-to-strong'' model pairs based on their specific needs. Depending on the type of the model pair (e.g., DreamShaper vs. SD1.5, good experts vs. bad experts in MoE), different effects of improvements can be achieved. We provide further application analysis in~\cref{sec:application} to demonstrate the broad applicability of W2SD.

The contributions can be summarized as follows.

First, we introduce the weak-to-strong concept into the inference enhancement of diffusion models, demonstrating that the gradient difference of estimated log probability densities between weak and strong diffusion models can approximate the difference between strong and ideal models, consequently bridging the gap between the learned and real data distribution.

Second, to estimate the weak-to-strong difference, we propose a novel inference framework called W2SD, with theoretical understanding that implicitly estimates the weak-to-strong difference through iterative reflection, effectively steering latent variables along sampling trajectories toward regions corresponding to the real data distribution.

Third, extensive experiments validate the effectiveness and broad applicability of W2SD across various generation tasks (e.g., image, video), architectures (e.g., UNet-based, DiT-based), and evaluation metrics. By defining various weak-to-strong model pairs, W2SD achieves diverse improvements effects, such as human preference, prompt adherence, and personalization.  Importantly, these improvements effects can be cumulative in parallel, further enhancing the quality of the generation. In efficiency evaluations, W2SD's improvements surpass its overhead, maintaining superior quality over the baseline under equal time constraints, suggesting its strong versatility and applicability.

\section{Preliminaries}
\label{sec:Preliminaries}
In this section, we present preliminaries about denoising and inversion operation in diffusion models~\bai{~\citep{songscore,ho2020denoising}}. Due to page limitations, we introduce the related work in~\cref{sec:related_work}.

Given the random Gaussian noise $z_{t}$, we denote the forward process of diffusion models as $x_{t}=x_{t-\Delta t}+\sigma^{t}\sqrt{\Delta t}z_{t}$, where $t\in [0,1]$, and $\sigma$ represents the predefined variance. We denote the ground truth density of real data distribution as $p_{0}^{\mathrm{gt}}$, After noise addition at time~$t$, the resulting density is represented as $p_{t}^{\mathrm{gt}}$.

Following~\citet{song2020denoising}, we can obtain the denoised results $x_{t- \Delta t}$ from noisy data $x_{t}$ through the process of an ordinary differential equation as
\begin{align}
    x_{t-\Delta t} &= \mathcal{M}(x_{t},t) \\
    &= x_{t} + \sigma^{2t}s_{\theta}(x_{t},t)\Delta t, \quad t \in [0,1].
  \label{eq:backward_ode}
\end{align}
where $s_{\theta}(\cdot,\cdot)$ represents the trained score network, utilized to predict the score at the time $t$. Similarly, we can invert~\cref{eq:backward_ode} to transform $x_{t-\Delta t}$ back to a new $\Tilde{x}_{t}$ as
\begin{align}
  \Tilde{x}_{t} &= \mathcal{M}_{\mathrm{inv}}(x_{t-\Delta t}, t),\\ 
  &=  x_{t-\Delta t} - \sigma^{2t}s_{\theta}(x_{t}, t)\Delta t, 
  \\
  &\approx x_{t-\Delta t} - \sigma^{2t}s_{\theta}(x_{t-\Delta t}, t)\Delta t,\quad t \in [0,1].
  \label{eq:inversion_ode_approx}
\end{align}

In practice, we often approximate the score value predicted at time $t$ with time $t-\Delta t$ along the inversion process, i.e., $s_{\theta}(x_{t},t) \approx s_{\theta}(x_{t-\Delta t},t)$ in~\cref{eq:inversion_ode_approx}. Given that the approximation error is negligible and the same socre netowrk $s_{\theta}$, $\mathcal{M}$ and $\mathcal{M}_{\mathrm{inv}}$ can be treated as mutually inverse mappings, thereby satisfying $\mathcal{M}_{\mathrm{inv}}(\mathcal{M}(x_{t},t),t)=x_{t}$. In~\cref{sec:error_exp}, we conduct a detailed analysis of the impact caused by this approximation error.

\section{Method}
\label{sec:detail_method}
In this section, we discuss the proposed Weak-to-Strong Diffusion and its visual explanation.

\subsection{Weak-to-Strong Diffusion}
\label{sec: theory}
In this subsection, we integrate the weak-to-strong concept into diffusion model inference, introduce the W2S algorithm, and establish its theoretical understanding.

\paragraph{The Difference of Estimated Density Gradients}
In diffusion models, the goal is to minimize the gradients difference of log probability densities between estimated results and ground truth. However, due to the real data distribution is inaccessible, this difference cannot be directly quantified. To tackle this issue, we consider models of varying capacities: a strong model (with its corresponding denoising process denoted as $\mathcal{M}^{\mathrm{s}}$ and the estimated density as $p^{\mathrm{s}}$) and a weak model (similarly, $\mathcal{M}^{\mathrm{w}}$ and $p^{\mathrm{w}}$). 

As shown in ~\cref{fig:motivation}, we define the weak-to-strong difference as $\Delta_{1}=\nabla \log{p^{\mathrm{s}}} - \nabla \log{p^{\mathrm{w}}}$ and the strong-to-ideal difference as $\Delta_{2}=\nabla \log{p^{\mathrm{gt}}} - \nabla \log{p^{\mathrm{s}}}$. By approximating $\Delta_{2}$ using the estimable $\Delta_{1}$, we indirectly reduce the gap between existing diffusion models and the ideal model, bringing the learned distribution closer to the real data distribution. In the next subsection, we introduce how to achieve this approximation from a reflective perspective.

\paragraph{W2SD} Consider a strong model $\mathcal{M}^{\mathrm{s}}$ and a weak model $\mathcal{M}^{\mathrm{w}}$, we can optimize the sampling trajectories using the reflection operator $\mathbf{\mathcal{M}^{\mathrm{w}}_{\mathrm{inv}}(\mathcal{M}^{\mathrm{s}}(\cdot))}$, as outlined in~\cref{algo:ReDiff}. Through the iterative integration of strong model denoising and weak model inversion, we achieve a step-by-step reflective process during sampling process, refining the latent variable $x_{t}$ into an improved $\Tilde{x}_{t}$.

\begin{algorithm}[h]
   \caption{W2SD}
   \label{algo:ReDiff}
\begin{algorithmic}
   \STATE {\bfseries Input:} Strong Model $\mathcal{M}^{\mathrm{s}}$, Weak Model $\mathcal{M}^{\mathrm{w}}$, Total Inference Steps: $T$, optimization steps: $\lambda$
   \STATE {\bfseries Output:} Clean Data $x_{0}$
   \STATE Sample Gaussian noise $x_{T}$
   \FOR{$t=T$ {\bfseries to} $1$}
   \IF{$t>T - \lambda$}
   \STATE \#W2SD with Reflection\\
   \STATE $\Tilde{x}_{t} = \mathcal{M}_{\mathrm{inv}}^{\mathrm{w}}(\mathcal{M}^{\mathrm{s}}(x_{t},t),t)$
   \ENDIF
   \STATE $x_{t-1} = M^{\mathrm{s}}(\Tilde{x}_{t},t)$
   \ENDFOR
\end{algorithmic}
\end{algorithm}

Importantly, the choice of $\mathcal{M}^{\mathrm{s}}$ and $\mathcal{M}^{\mathrm{w}}$ significantly affects the direction of improvements effects. We summarize some promising weak-to-strong model pairs in~\cref{tab:meta_appendix} of~\cref{sec:exp_dataset}. And in~\cref{sec:application} we present extensive application analyses and experiments, highlighting the powerful capabilities and flexibility of the proposed framework.

\begin{figure}[t]
    \centering
\includegraphics[width=1\linewidth]{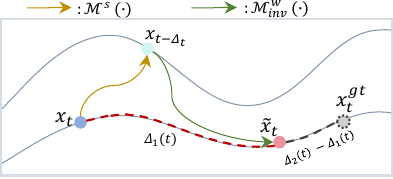}
\vspace{-0.25cm}
    \caption{Visualizing the effectiveness of W2SD. When the weak-to-strong difference closely approximates the strong-to-ideal difference (e.g., $\Delta_{2}(t)-\Delta_{1}(t)$ is small), the refined latent variable $\Tilde{x}_{t}$ converges to the ideal latent variable $x_{t}^{\mathrm{gt}}$.}
    \label{fig:w2sd_effect}
\end{figure}

In~\cref{theorem:1}, we demonstrate that W2SD refines latent variable $x_{t}$ toward the direction defined by the estimated weak-to-strong difference $\Delta_{1}(t)$. As shown in~\cref{fig:w2sd_effect}, when the weak-to-strong difference closely approximates the strong-to-ideal difference (i.e., $\Delta_{2}(t) - \Delta_{1}(t)$ is small), the reflection mechanism of W2SD drives $x_{t}$  closer to the ideal $x_{t}^{\mathrm{gt}}$. The visualization analysis in~\cref{sec:visual_explan} validates the correctness of our theory, and we provide a detailed proof in~\cref{proof:1}.

\begin{theorem}[Theoretical Understanding of W2SD]
\label{theorem:1}
 Suppose $x_{t}$ is the latent variable at time $t$, let $p_{t}^{\mathrm{s}}$ and $p_{t}^{\mathrm{w}}$ denote the probability density estimates derived from $\mathcal{M}^{\mathrm{s}}$ and $\mathcal{M}^{\mathrm{w}}$. The reflective operator $\mathcal{M}^{\mathrm{w}}_{\mathrm{inv}}(\mathcal{M}^{\mathrm{s}}(\cdot))$ refines $x_{t}$ to $\Tilde{x}_{t}$ as

\begin{equation}
\label{eq:theorem_1_eq}
    \Tilde{x}_{t} = x_{t} + \sigma^{2t}\Delta t ( \nabla_{x_{t}} \log{p_{t}^{\mathrm{s}}(x_{t})} - \nabla_{x_{t}} \log{p_{t}^{\mathrm{w}} (x_{t})} ),
\end{equation}
where $\Delta_{1} (t) = \nabla_{x_{t}} \log{p_{t}^{\mathrm{s}}(x_{t})} - \nabla_{x_{t}} \log{p_{t}^{\mathrm{w}} (x_{t})}$ means the weak-to-strong difference between $\mathcal{M}^{\mathrm{s}}$ and $\mathcal{M}^{\mathrm{w}}$ at time $t$.
\end{theorem}

It is noted that~\citet{karras2024guiding} proposed Auto-guidance, which employs a simplistic interpolation approach using a degraded model version to refine latent variables. While they shares a similar high-level concept, W2SD introduces fundamentally distinct mechanisms, delivering significantly stronger performance and broader applicability. Comprehensive quantitative and qualitative analyses are provided in~\cref{sec:auto-guidance}.

\subsection{Visualization and Explanation in Various Settings}
\label{sec:visual_explan}
In this subsection, we validate the theory presented in~\cref{sec: theory} using both synthetic Gaussian mixture data and real-world image data, providing intuitive visual evidence.

\paragraph{1-D Gaussian Mixture Data}
We begin by analyzing the 1-D Gaussian data scenario. In~\cref{fig:motivation}, the diffusion model is designed to generate data with two distinct peaks at ``-4'' and ``4''. Adjusting the proportion of these two peaks in the training dataset, we obtain $\mathcal{M}^{\mathrm{s}}$ and $\mathcal{M}^{\mathrm{w}}$. Although both models effectively generate samples near the right peak, their performance differs significantly for the left peak.

\begin{figure}[t]
    \centering
    \includegraphics[width=1.0\linewidth]{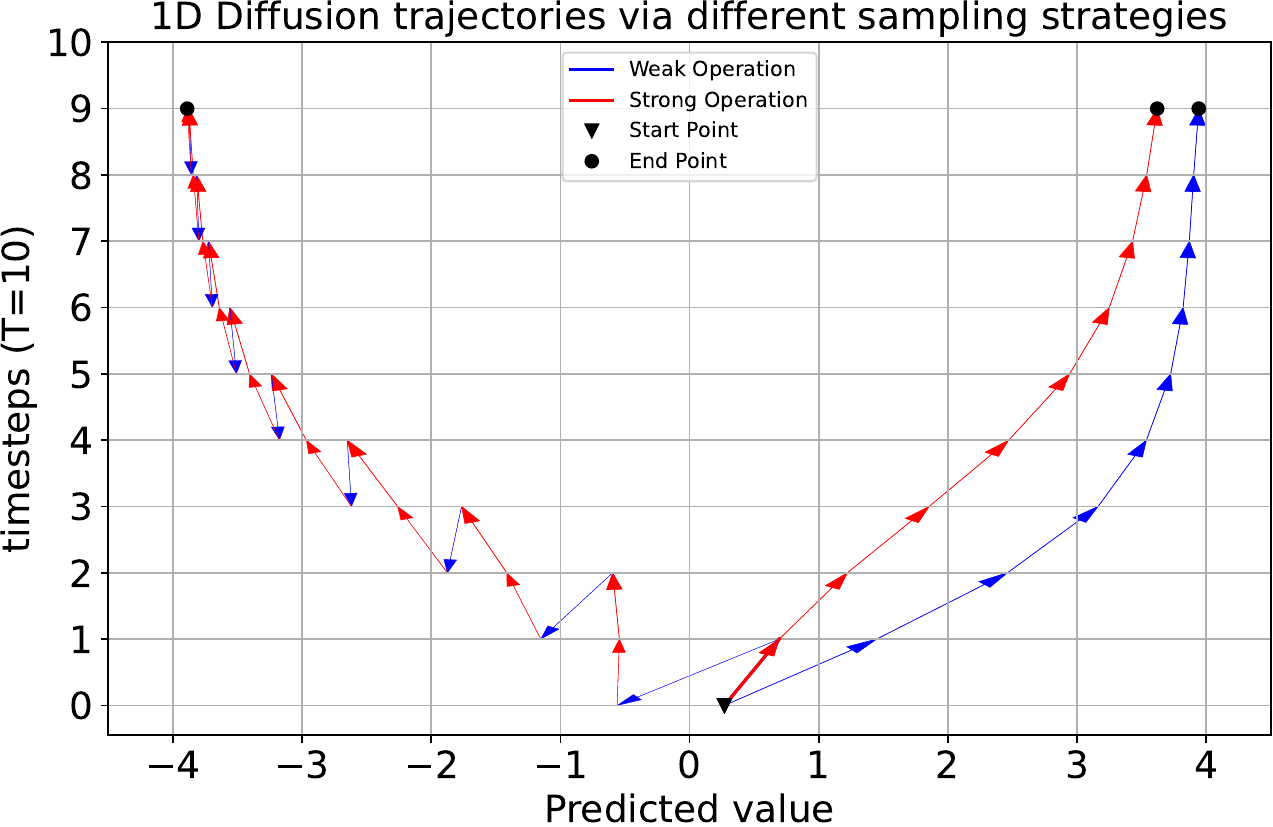}
    \vspace{-0.30cm}
    \caption{Denoising trajectories across different settings (1-D Gauss). The weak model (blue) generates only right-peak data due to missing left-peak training samples, while the strong model (red) produces data between both peaks. W2SD balances the distribution by leveraging the reflective operator $\mathcal{M}_{\mathrm{inv}}^{\mathrm{w}}(\mathcal{M}^{\mathrm{s}}(\cdot))$. Additional examples are provided in ~\cref{fig:1d-more-case}.}
    \label{fig:1d-gauss}
\end{figure}

In~\cref{fig:1d-gauss}, we visualize the denoising trajectories under three different settings (weak model, strong model, W2SD). Through the reflective operator $\mathbf{M}_{\mathrm{inv}}^{\mathrm{w}}(\mathcal{M}^{\mathrm{s}}(\cdot))$, the latent variable is progressively drawn closer to the left peak. In contrast, for both strong and weak models, the generated samples are predominantly concentrated around the right peak.

\begin{figure}[t]
    \centering
    \includegraphics[width=1.0\linewidth]{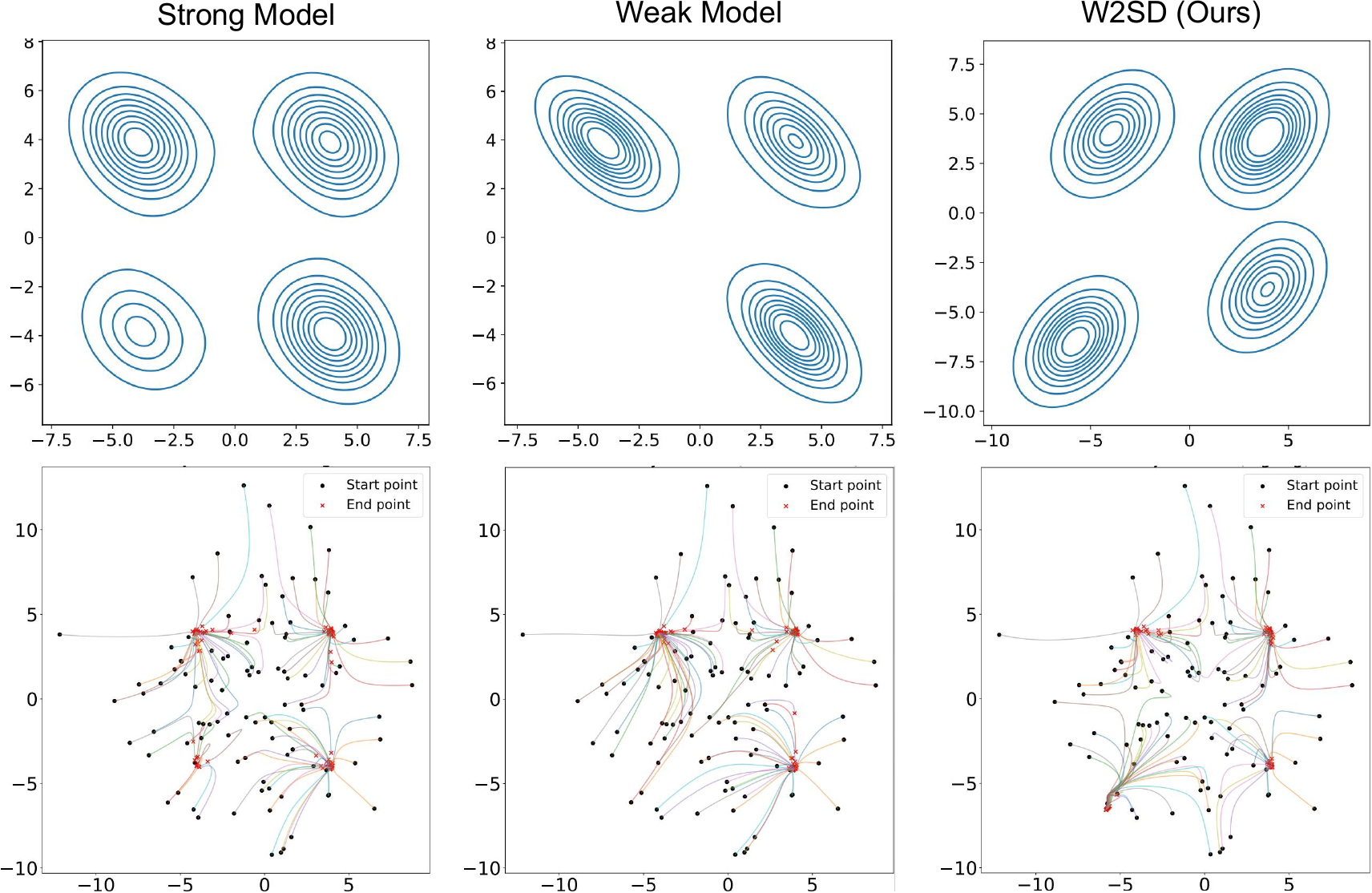}
    \vspace{-0.25cm}
    \caption{Probability contour plot and denoising trajectories across different settings (2-D Gauss). W2SD balances the learned distribution, bringing it closer to the real data distribution}.
    \label{fig:2d-gauss}
\end{figure}

\paragraph{2-D Gaussian Mixture Data}
We also visualize the mechanism of W2SD on 2D scenario. In~\cref{fig:2d-gauss} (first row), we modulate the proportion of the training data to obtain $\mathcal{M}^{\mathrm{s}}$ and $\mathcal{M}^{\mathrm{w}}$. W2SD balances the distribution by exploiting the discrepancy between $\mathcal{M}^{\mathrm{s}}$ and $\mathcal{M}^{\mathrm{w}}$ (the region in the bottom-left corner, denoted as $\Delta_{1}$) to approximate the unattainable strong-to-ideal difference $\Delta_{2}$, and boost the chances of sampling toward the bottom-left region. In~\cref{fig:2d-gauss} (second row), we visualize the denoising trajectories under different settings, further validating the effectiveness of W2SD.

\paragraph{Real Image Data}
\begin{figure}[t]
    \centering
    \includegraphics[width=1.0\linewidth]{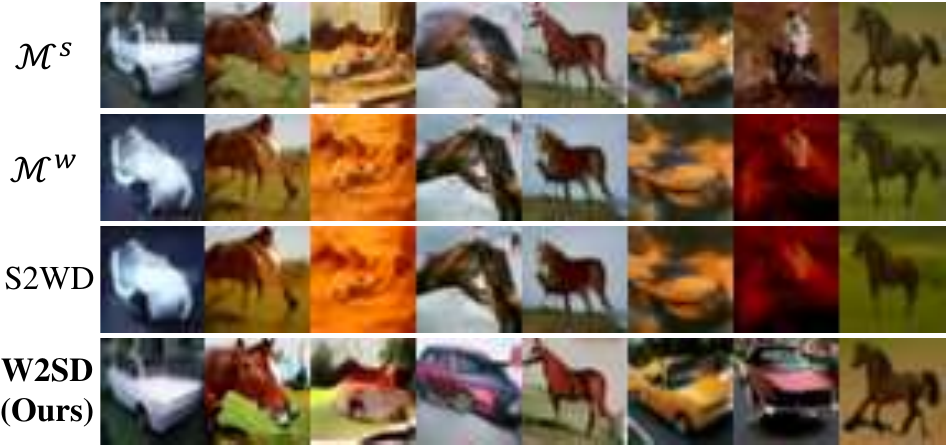}
    \vspace{-0.25cm}
    \caption{Qualitative results of W2SD based on dataset differences (CIFAR-10). Our method enhances the probability of generating ``cars'' and promote a more balanced generation distribution.}
    \label{fig:case_dataset_gap_visual}
\end{figure}
Furthermore, we investigate the performance of W2SD on real image data. For ease of analysis, we select two classes from CIFAR-10~\citep{krizhevsky2009learning}: \textbf{car} and \textbf{horse}. Specifically, we train the weak and strong diffusion models on distinct datasets. For $\mathcal{M}^{\mathrm{s}}$, the dataset consists of 5,000 horse images and 2,500 car images. For $\mathcal{M}^{\mathrm{w}}$, it comprises 5,000 horse images and 500 car images. In this scenario, due to the imbalance in training dataset, both $\mathcal{M}^{\mathrm{w}}$ and $\mathcal{M}^{\mathrm{s}}$ are more inclined to generate horses. Moreover, since $\mathcal{M}^{\mathrm{w}}$ is trained on a limited number of car images, it rarely generates cars.

\begin{figure}[t]
    \centering
    \includegraphics[width=1.0\linewidth]{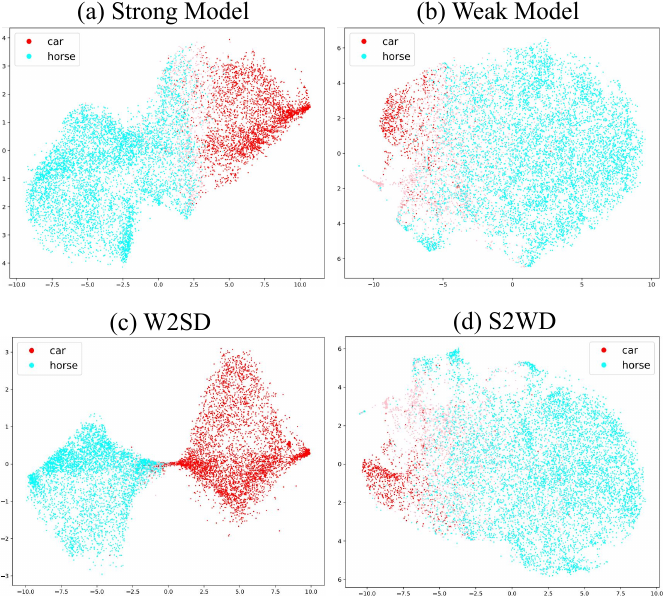}
    \vspace{-0.3cm}
    \caption{The CLIP feature corresponding to the generated image (32$\times$32$\times$3) is projected into a 2D space. W2SD effectively disentangles the
representations of ``car'' and ``horse'' in the 2D space. \textbf{(a)} $\mathcal{M}^{\mathrm{s}}$ demonstrates the ability to generate cars; \textbf{(b)} $\mathcal{M}^{\mathrm{w}}$ can hardly generate cars; \textbf{(c)} W2SD balances the generation distribution, increasing the likelihood of generating cars; \textbf{(d)} S2WD (i.e., $\mathcal{M}_{inv}^{s}(\mathcal{M}^{w}(\cdot))$) exacerbates the imbalance in data generation.}
    \label{fig:case_dataset_gap_map_2d}
\end{figure}

In~\cref{fig:case_dataset_gap_visual}, we note that W2SD can help balance the image categories, enhancing inference process to increase the probability of generating cars. In~\cref{fig:case_dataset_gap_map_2d}, we also perform a t-SNE dimensionality reduction~\citep{van2008visualizing} on the CLIP features of the generated images. W2SD effectively disentangles the representations of ``car'' and ``horse'' in the 2D space. Notably, the ratio of ``horse'' to ``car'' under W2SD approaches 1:1. In contrast, applying the negative reflection operator $\mathcal{M}_{\mathrm{inv}}^{\mathrm{s}}(\mathcal{M}^{\mathrm{w}}(\cdot))$ worsens the data imbalance, validating the effectiveness of our method.

\section{Empirical Analysis}
\label{sec:application}
In this section, we justify our design choices in ~\cref{sec:detail_method} and illustrate the wide applicability of W2SD across diverse combinations of $\mathcal{M}^{\mathrm{s}}$ and $\mathcal{M}^{\mathrm{w}}$.

Due to page limitations, 
here we validate the effectiveness of W2SD using Pick-a-Pic Dataset~\citep{kirstain2023pick}. In~\cref{sec:exp_setting}, we provide a detailed description of the experimental settings. And in~\cref{sec:exp_res}, we conduct more extensive quantitative and qualitative results across diverse benchmarks (e.g., Drawbench~\citep{saharia2022photorealistic}) and modalities (e.g., video generation), demonstrating the broad applicability of our method.

\begin{figure*}[t]
    \centering
    \includegraphics[width=0.95\linewidth]{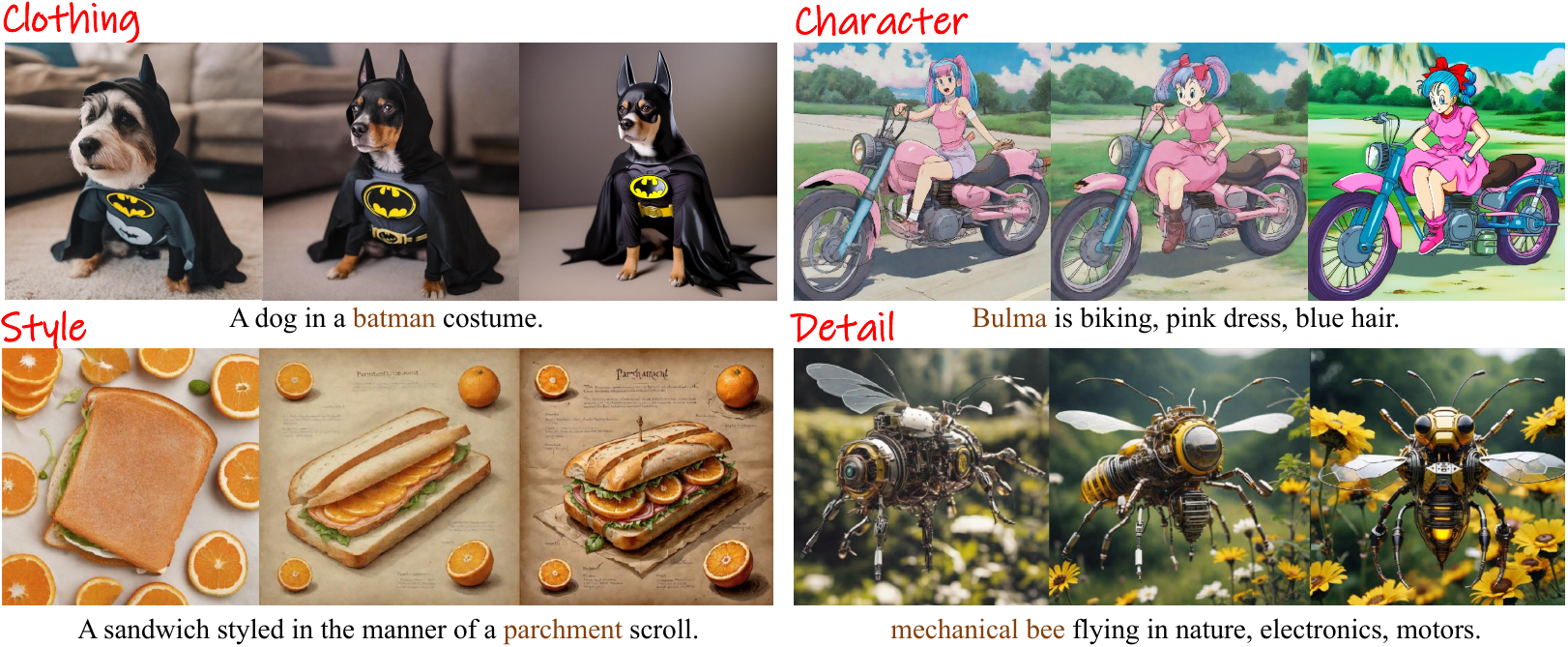}
    \caption{Qualitative comparisons with weak model (\textbf{left}), strong model (\textbf{middle}) and W2SD based on weight difference (\textbf{right}). Our method utilizes the differences between chosen strong and weak models (e.g., high-detail LoRA vs. standard model) to deliver improvements in various dimensions, including style, character, clothing, and beyond. We provide more qualitative results in~\cref{sec:exp_res_quali}.}
    \label{fig:case_weight_gap}
\end{figure*}

\subsection{Weight Difference}
The capability differences between models can be directly captured by their parameter weights. And W2SD leverages this weight difference to empirically estimate the weak-to-strong difference, enabling effective reflective operations.

\paragraph{Full Parameter Fine-tuning} We first select the full parameter fine-tuned models (e.g., DreamShaper, Juggernaut-XL) that align more closely with human preferences as $\mathcal{M}^{\mathrm{s}}$, and the corresponding standard models as $\mathcal{M}^{\mathrm{w}}$ (e.g., SD1.5~\citep{rombach2022high}, SDXL~\citep{podellsdxl}). We evaluate our method with various metrics, as shown in~\cref{tab:full_finetune}, W2SD based on weight difference shows significant improvements in human preference metrics such as HPS v2~\citep{wu2023human} and PickScore~\citep{kirstain2023pick}. 

\begin{table}[t]
\caption{Quantitative results of W2SD based on a full parameter fine-tuning strategy. Our method generates results better aligned with human preferences. Datasets: Pick-a-Pic.}
\label{tab:full_finetune}
\begin{center}
\begin{small}
\resizebox{0.5\textwidth}{!}{
\begin{tabular}{c|cccc}
\toprule
Method & HPS v2 $\uparrow$ & AES $\uparrow$ & PickScore $\uparrow$& MPS $\uparrow$\\
\midrule
SD1.5 & 24.9558 & 5.5003 & 20.1368 & - \\
\midrule
DreamShaper & 30.1477 & 6.1155 & 21.5035 & 46.8705  \\
\textbf{W2SD} & \textbf{30.4924} & \textbf{6.2478} & \textbf{21.5727} & \textbf{53.1304} \\
\midrule
\midrule
SDXL & 29.8701 & \textbf{6.0939} & 21.6487 & -  \\
\midrule
Juggernaut-XL & 31.6412 & 5.9790 & 22.1903 & 45.7397 \\
\textbf{W2SD}  & \textbf{32.0992} & 6.0712 & \textbf{22.2434} & \textbf{54.2634} \\
\bottomrule
\end{tabular}
}
\end{small}
\end{center}
\end{table}

\begin{table}[t]
\caption{Quantitative results of W2SD based on personalized LoRA model. Here, the weight difference between $\mathcal{M}^{w}$ (SD1.5) and $\mathcal{M}^{s}$ (SD1.5 +LoRA) biases the generated results towards a more personalized direction.}

\label{tab:ckpt_gap_personal}
\begin{center}
\begin{small}
\resizebox{0.5\textwidth}{!}{
\begin{tabular}{c|cccc}
\toprule
Method & DINO $\uparrow$& CLIP-I $\uparrow$& CLIP-T $\uparrow$& \\
\midrule
SD1.5 & 27.47 & 52.08 & 20.14 \\
Personalized LoRA & 48.03 & 64.37 & 25.99 \\
\textbf{W2SD} & \textbf{51.58} & \textbf{68.04} & \textbf{27.66} \\
\bottomrule
\end{tabular}
}
\end{small}
\end{center}
\end{table}

\paragraph{LoRA Mechanism} Furthermore, W2SD is also applicable to efficiently fine-tuned model. we select the personalized models derived through the LoRA mechanism~\citep{hulora} as $\mathcal{M}^{\mathrm{s}}$, and employ W2SD to attain a more robust and customized personalization effect. We test 20 LoRA checkpoints across object, person, animal, and style categories, with details in~\cref{sec:exp_dataset}. ~\cref{tab:ckpt_gap_personal} demonstrates that W2SD results in significant improvements across multiple personalization metrics (e.g., Clip-T and Clip-I). And we present the qualitative results in~\cref{fig:case_weight_gap}, with more visual cases provided in~\cref{sec:exp_res_quali}.

\begin{table}[t]
\caption{Quantitative results of W2SD based on MoE Mechanism. Datasets: ImageNet 50K.}
\label{tab:moe}
\begin{center}
\begin{small}
\resizebox{0.5\textwidth}{!}{
\begin{tabular}{c|cccc}
\toprule
Method & IS $\uparrow$ & FiD $\downarrow$ & AES $\uparrow$ & HPS v2 $\uparrow$\\
\midrule
DiT-MoE-S & 45.4437 & 15.1032 & 4.4755 & 20.0486\\
\textbf{W2SD} & \textbf{55.5341} & \textbf{9.1001} & \textbf{4.5053}& \textbf{22.3225}\\
\bottomrule
\end{tabular}
}
\end{small}
\end{center}
\end{table}

\begin{figure}[h!]
    \centering
    \includegraphics[width=1\linewidth]{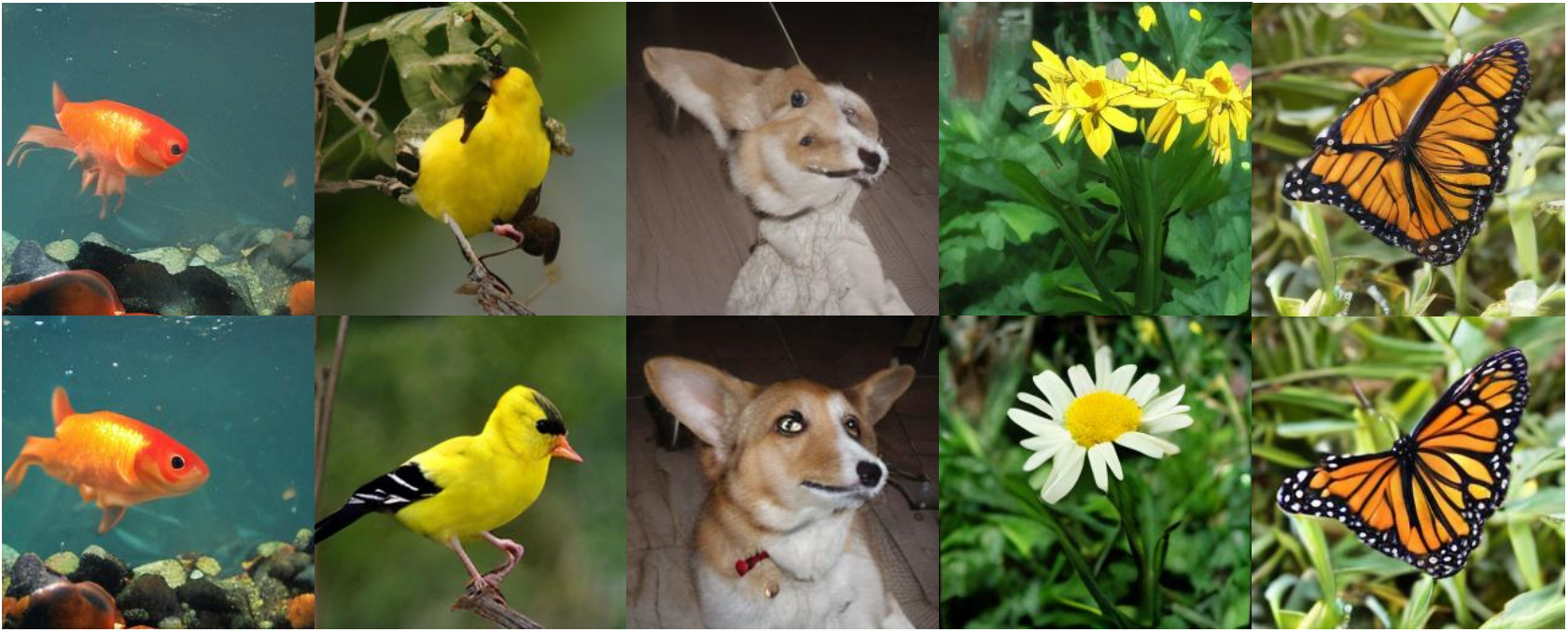}
    \vspace{-0.25cm}
    \caption{Quantitative Results of W2SD Based on the MoE Mechanism. The \textbf{first row} shows the results for DiT-MoE-S, while the \textbf{second row} presents W2SD. W2SD achieves significant improvements, even with small models featuring 71M activated parameters.}
    \label{fig:moe_case}
\end{figure}

\paragraph{MoE Mechanism}
We also note that a weak-to-strong difference can be induced by controlling the expert selection strategy within the MoE mechanism. Specifically, we focus on DiT-MoE~\citep{fei2024scaling}, a novel architecture that integrates multiple experts and selects the two highest-performing experts during each denoising step to optimize generation quality. Based on this framework, we use DiT-MoE-S as $\mathcal{M}^{\mathrm{s}}$ and define $\mathcal{M}^{\mathrm{w}}$ as the two lowest-performing experts in each denoising step, thereby establishing a quantifiable weight difference.

In~\cref{tab:moe}, we show that W2SD reduces FiD~\citep{Seitzer2020FID} from 15.1032 to \textbf{9.1001}, aligning the learned distribution closer to the real data distribution. Additionally, as shown in~\cref{fig:moe_case}, although the limited capacity of MoE-DiT-S (with only 71M active parameters) often results in image degradation and distortion (the first row in~\cref{fig:moe_case}), the application of W2SD significantly enhances the image quality.

\subsection{Condition Difference}
Here we demonstrate that the weak-to-strong difference applies not only across different weights but also within the same diffusion model under the weak-to-strong conditions.

\begin{table}[t]
\caption{Quantitative results of W2SD based on guidance difference. Model: SDXL. Datasets: Pick-a-Pic.}
\label{tab:cfg_gap}
\begin{center}
\begin{small}
\resizebox{0.5\textwidth}{!}{
\begin{tabular}{c|cccc}
\toprule
Method & HPS v2 $\uparrow$& AES $\uparrow$& PickScore $\uparrow$& MPS $\uparrow$\\
\midrule
SD1.5 & 24.9558 & 5.5003 & 20.1368 & 42.1101 \\
\textbf{W2SD} & \textbf{25.5069} & \textbf{5.5073} & \textbf{20.2443} & \textbf{57.8903} \\
\midrule
SDXL & 29.8701 & 6.0939 & 21.6487 & 43.9425\\
\textbf{W2SD} & \textbf{31.2020} & \textbf{6.0970} & \textbf{21.7980} & \textbf{56.0608}\\
\bottomrule
\end{tabular}
}
\end{small}
\end{center}
\end{table}

\paragraph{Guidance Scale} The classifier-free guidance mechanism~\citep{ho2021classifier} extrapolates between conditional and unconditional noise predictions, adjusting the guidance scale to control the semantic attributes of the generated images. To apply W2SD, diffusion models under high guidance scale can be viewed as $\mathcal{M}^{\mathrm{s}}$, those under low or even negative guidance scale can be treated as $\mathcal{M}^{\mathrm{w}}$. This guidance scale discrepancy constructs a weak-to-strong difference that aligns the learned generated distribution more closely with the semantic conditions of the prompt.

We note that Z-Sampling~\citep{bai2024zigzag} is a specific instance of W2SD under the guidance difference. In~\cref{tab:cfg_gap}, our method significantly enhances human preference (e.g., HPS v2) and aesthetic characteristics (e.g., AES) in mainstream models such as SDXL and SD1.5.

\paragraph{Semantics of Prompts}
Aside from adjusting the guidance scale, we demonstrate that the semantic differences within the condition prompts themselves can also be leveraged to enhance generation quality through the reflection process. 

\begin{table}[t]
\caption{Quantitative results of W2SD based on semantic differences between prompts. Model: SDXL. Datasets: GenEval.}
\label{tab:prompt_gap}
\begin{center}
\begin{small}
\resizebox{0.48\textwidth}{!}{
\begin{tabular}{c|cccc}
\toprule
Prompt Type & HPS v2 $\uparrow$ & AES $\uparrow$ & PickScore $\uparrow$ & MPS $\uparrow$\\
\midrule
Raw Prompt & 25.3897 & 5.4454 & 20.7144 & -\\
Refined Prompt & 28.5698 & 5.7714 & 21.6350 & 45.7719\\
\textbf{W2SD} & \textbf{29.4023} & \textbf{5.8812} & \textbf{21.8053} & \textbf{54.2275}\\
\bottomrule
\end{tabular}
}
\end{small}
\end{center}
\vspace{-0.2cm}
\end{table}

\begin{table}[t!]
\caption{The improvements effects from different model differences can be cumulative. Datasets: Pick-a-Pic.}
\label{tab:cumulative}
\begin{center}
\begin{small}
\resizebox{0.45\textwidth}{!}{
\begin{tabular}{cc|ccr}
\toprule
Weight Difference & Condition Difference & HPS v2 $\uparrow$ & Winning Rate $\uparrow$\\
\midrule
$\times$ & $\times$ & 31.6412 & -\\
$\times$ & $\checkmark$ & 32.8217 & 84\%\\
$\checkmark$ & $\times$ & 32.0992 & 76\%\\
$\checkmark$ & $\checkmark$ & \textbf{32.9623} & \textbf{90\%}\\
\bottomrule
\end{tabular}
}
\end{small}
\end{center}
\end{table}

Specifically, we select a total of 533 text prompts from the GenEval~\citep{ghosh2024geneval} for $\mathcal{M}^{\mathrm{w}}$, with an average length ranging from 5 to 6 words and minimal contextual complexity. Additionally, we leverage LLM (here we use Qwen-Plus~\citep{yang2024qwen2}) to enrich and refine these prompts with additional details for $\mathcal{M}_{\mathrm{s}}$. resulting in weak-to-strong prompt pairs.

We report the quantitative results in~\cref{tab:prompt_gap}, which relies solely on the semantic differences between prompts, achieves improvements across all metrics. And in~\cref{fig:case_prompt_gap} of~\cref{sec:exp_res_quali} we illustrate that W2SD is capable of more precisely capturing the intricate details based on the differences in prompt pairs.

\subsection{Sampling Pipeline Difference}
Extensive works~\citep{si2023freeu, chefer2023attendandexcite, zhang2023adding} have been devoted to design advanced inference pipelines to improve the quality of diffusion generation results. We demonstrate that W2SD can enhance the inference process by leveraging the difference derived from these powerful existing pipelines. Here we define those enhanced sampling methods as $\mathcal{M}^{\mathrm{s}}$, while $\mathcal{M}^{\mathrm{w}}$ represents the standard sampling~\citep{song2020denoising, lu2022dpm}. 

\begin{figure}[h]
    \centering
    \includegraphics[width=0.95\linewidth]{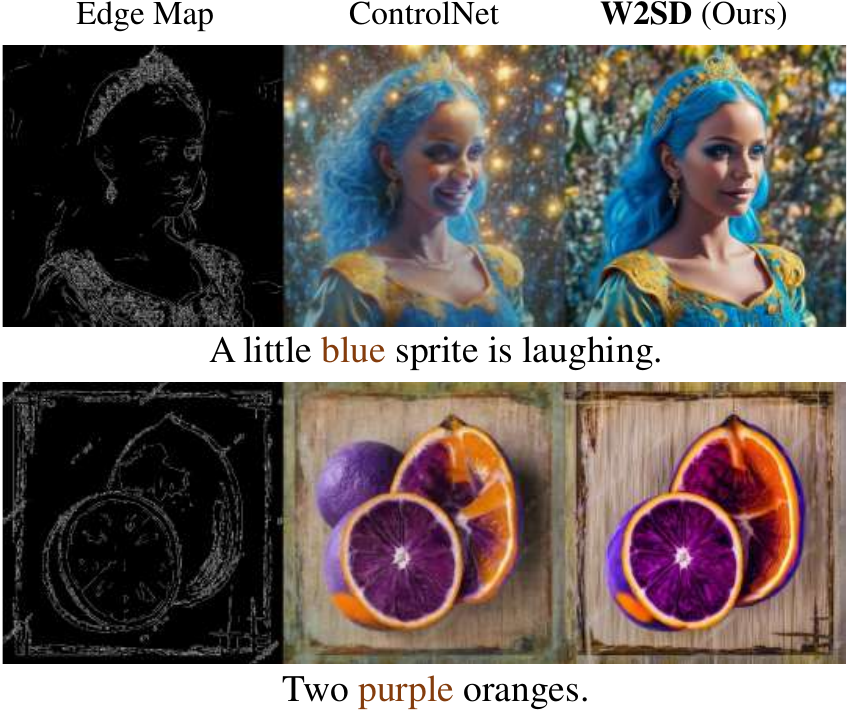}
    \vspace{-0.1cm}
    \caption{Qualitative results of W2SD based on pipeline difference. We set ControlNet as $\mathcal{M}^{\mathrm{s}}$, DDIM as $\mathcal{M}^{\mathrm{w}}$. W2SD improves alignment with reference images.}
    \label{fig:main_case_controlnet}
\end{figure}

We first select ControlNet~\citep{zhang2023adding} as $\mathcal{M}^{\mathrm{s}}$, which incorporates additional network structures into the pipeline. By utilizing reference images (e.g., edge maps), it facilitates controllable image generation. As shown in~\cref{fig:main_case_controlnet}, the images generated by W2Sd exhibit a stronger alignment with the provided edge maps. We also present visualization cases of other pipelines (e.g., Ip-adapter~\citep{ye2023ip}) as strong models in~\cref{fig:ip-adapter} of~\cref{sec:exp_res_quali}.

\subsection{Cumulative Effects of Different Model Differences}
Finally, it is important to note that the improvements effects of W2SD, derived from different type of differences, can be even cumulative. Here we apply the weight difference and the condition difference simultaneously, where $\mathcal{M}^{\mathrm{s}}$ represents the fine-tuned model Juggernaut-XL with high guidance scale at 5.5, while $\mathcal{M}^{\mathrm{w}}$ represents the standard model SDXL with low guidance scale at zero. As shown in~\cref{tab:cumulative}, the combination of guidance difference and condition difference leads to a substantial improvement in Pick-a-Pic Dataset compared to $\mathcal{M}^{\mathrm{s}}$, achieving a winning rate of up to 90\% on HPS v2.

\section{More Detailed Analysis}
In this section, we perform more detailed studies on W2SD, to further validate the effectiveness of our theory.

\subsection{The Magnitude of Weak-to-Strong Difference}
\label{sec: abla_diff}
In this subsection, we explore how the magnitude of weak-to-strong difference affects the improvements effects. 

\begin{figure}[t!]
    \centering
    \includegraphics[width=0.95\linewidth]{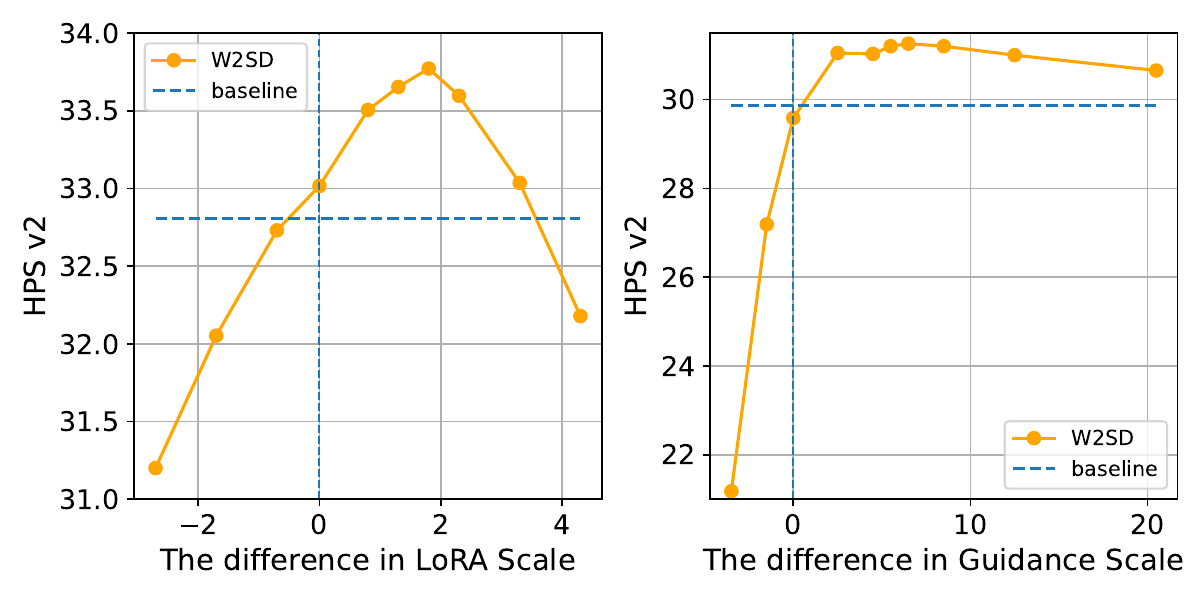}
    \caption{The magnitude of weak-to-strong difference is a key factor impacting the
effects of improvements. The horizontal axis shows the magnitude of the weak-to-strong difference, while the vertical axis
shows the average HPS v2 on the Pick-a-Pic. When $\mathcal{M}^{\mathrm{s}}$ is weaker than $\mathcal{M}^{\mathrm{w}}$, W2SD results in negative gains.}
    \label{fig:gap_abla}
\end{figure}

\begin{figure}[t]
    \centering
    \includegraphics[width=0.95\linewidth]{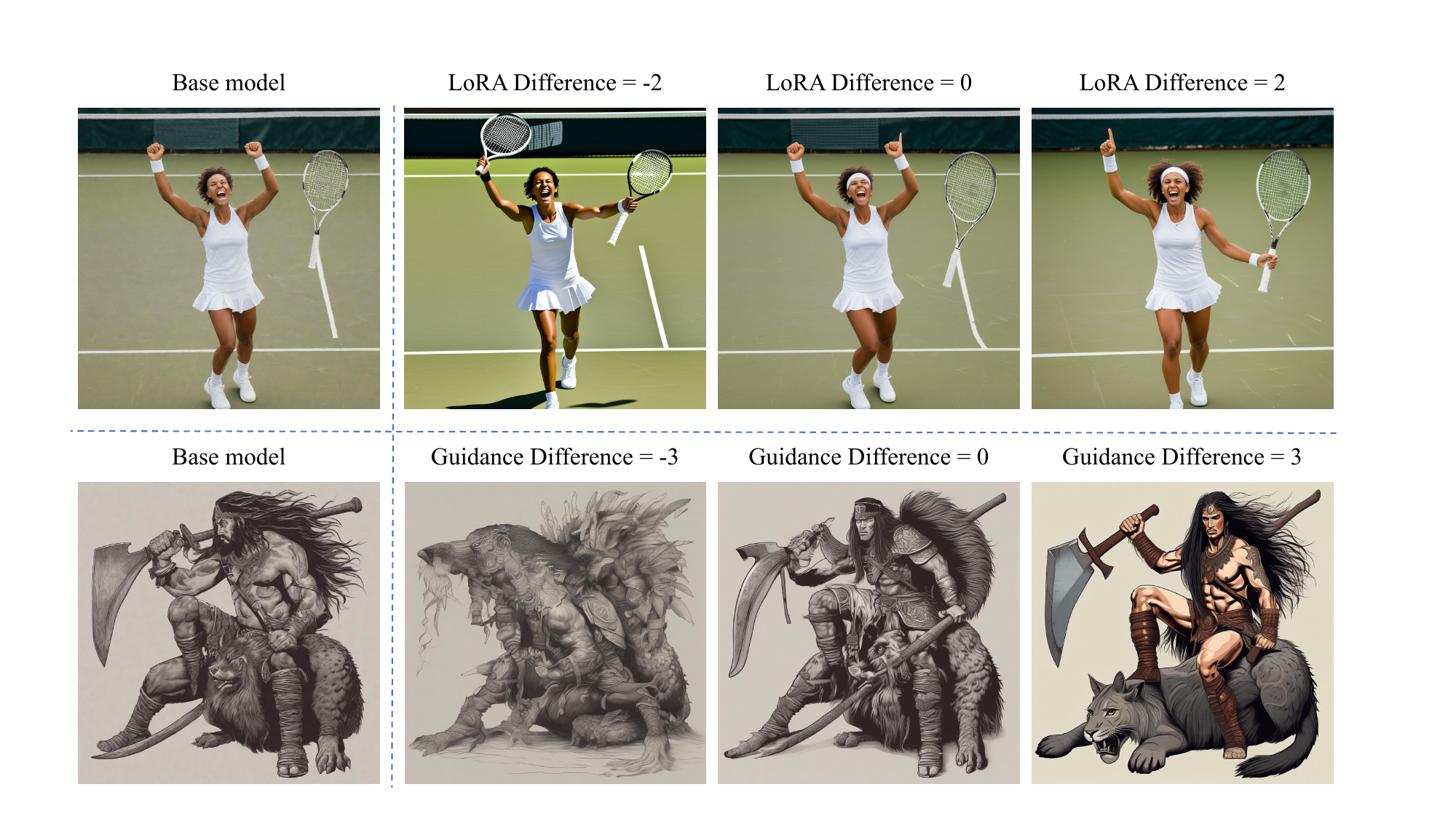}
    \caption{When the weak-to-strong difference is greater than 0, W2SD yields positive gains. When it equals 0, the process degenerates into standard sampling. When it is less than 0, negative gains occurs, resulting in poor image quality.}
    \label{fig:gap_abla_case}
    \vspace{-0.5cm}
\end{figure}

To quantify the difference between $\mathcal{M}^{\mathrm{s}}$ and $\mathcal{M}^{\mathrm{w}}$, we first consider the LoRA-based W2SD approach. We fix the LoRA scale of $\mathcal{M}^{\mathrm{s}}$ at 0.8 and adjust the LoRA scale of $\mathcal{M}^{\mathrm{w}}$ (e.g., from -3.5 to 3.5) to observe its impact on the effects of improvements. As shown in~\cref{fig:gap_abla} (left), when the LoRA scale of $\mathcal{M}^{\mathrm{s}}$ exceeds that of $\mathcal{M}^{\mathrm{w}}$, W2SD improves performance. However, when $\mathcal{M}^{\mathrm{s}}$ is weaker than $\mathcal{M}^{\mathrm{w}}$ (i.e., the LoRA scale difference is negative), the improvements effects diminish or even result in the negative gains.

In addition, as shown in~\cref{fig:gap_abla} (right), we also quantify the weak-to-strong difference based on guidance scale. By fixing the guidance scale of  $\mathcal{M}^{\mathrm{s}}$ at 5.5 and adjusting that of $\mathcal{M}^{\mathrm{w}}$ (e.g., from -10 to 15), we observe similar phenomena. In~\cref{fig:gap_abla_case}, we present a qualitative analysis showing that the magnitude of the weak-to-strong difference is a key factor that influences the quality of generation results.

\subsection{Time Efficiency Comparison} 
\begin{figure}[t]
    \centering
    \includegraphics[width=0.8\linewidth]{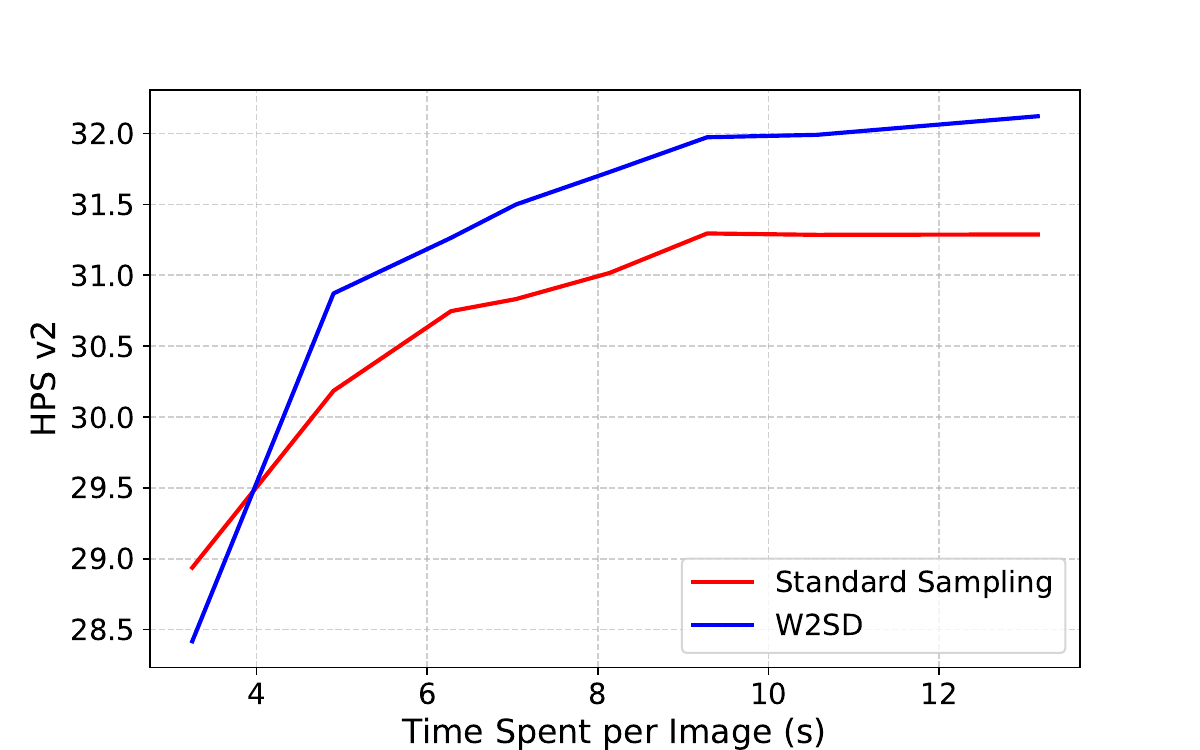}
    \caption{W2SD outperforms standard sampling with identical time costs. The horizontal axis denotes the average generation time per image, the vertical axis represents the HPS v2 on Pick-a-Pic.}
    \label{fig:time_abla}
    \vspace{-0.3cm}
\end{figure}

Assuming standard sampling and W2SD require $T_{\mathrm{std}}$ and $T_{\mathrm{w2s}}$ denoising steps, respectively, the score predictions needed are $T_{\mathrm{std}}$ and $T_{\mathrm{w2s}}+2\lambda$. To ensure a fair comparison, we set $T_{\mathrm{w2s}} = \lfloor \frac{1}{2} T_{\mathrm{std}} \rfloor$ and $\lambda = \lfloor \frac{1}{2} T_{\mathrm{w2s}} \rfloor$, matching the runtime for generating the same image. In this setting, as shown in~\cref{fig:time_abla}, W2SD consistently outperforms standard sampling, demonstrating that the gains from reflection operations far outweigh the additional computational cost, validating the time efficiency of our method.

\subsection{Appendix Overview}
Due to page limitations, further details are provided in the appendix:~\cref{sec:error_exp} analysis the impact of inversion approximation errors on performance improvement,~\cref{sec:other_methods} analyzes the connections between W2SD, Re-Sampling~\citep{lugmayr2022repaint}, and other inference methods.

\section{Conclusion}
To the best of our knowledge, this work is the first to systematically integrate the weak-to-strong concept into the inference enhancement of diffusion models via a reflection mechanism. We theoretically and empirically understand that the estimated weak-to-strong difference can effectively bridge the strong-to-ideal difference, enhancing the alignment between the learned distributions from existing diffusion models and the real data distribution. Building on this concept, we propose W2SD, which utilizes the estimated difference in density gradients to optimize sampling trajectories via reflective operations. W2SD demonstrates its effectiveness as a general-purpose framework through its cumulative performance improvements, flexible definition of weak-to-strong model pairs, and efficient performance gains with minimal computational overhead. We hope this work inspires interest in diffusion model scaling laws, advances sampling methods, and deepens understanding of probabilistic modeling in generative models.


\section*{Impact Statement}
This work introduces W2SD, a novel framework designed to enhance the inference process of diffusion models. The data and models utilized in our work are released under open-source licenses and sourced from open platforms. While our work may have various societal implications, it does not introduce additional ethnic concerns compared to existing standard sampling methods. As such, none which we feel must be specifically highlighted here.

\bibliography{ref}
\bibliographystyle{icml2025}

\newpage
\appendix
\onecolumn

\section{Related Work}
\label{sec:related_work}
In this section, we review existing works relevant to W2SD.

\paragraph{Weak-to-Strong Mechanism} The concept of improving weak models into strong models originates from the AdaBoost~\citep{hogsgaard2023adaboost}, which constructs a more accurate classifier by aggregating multiple weak classifiers. Building upon this, ~\citet{green2022optimal,hogsgaard2023adaboost} introduced a provably optimal weak-to-strong learner, establishing a robust theoretical foundation for this weak-to strong paradigm. In the field of LLMs, several studies~\citep{chenself,burns2023weak} have utilized weak models as supervisory signals to facilitate the alignment of LLMs. This paradigm of weak-to-strong generation during training has similarly been investigated in the context of diffusion model training~\citep{chen2025pixart}. And~\citet{sugiyama2013density} recommended directly estimating the density difference between weak and strong models instead of separate estimations. And~\citet{karras2024guiding} examines the mechanism of CFG and achieves improved results through interpolation-based perturbation. In contrast, W2SD utilizes the reflection by leveraging the gap between denoising and inversion to steer generation towards user-defined directions. And we note that W2SD generalizes the concept of ideal model, with experiments across human preference, personalization, fidelity and so on, confirming its wide applicability.

\paragraph{Diffusion Inference Enhancement}
The study of inference scaling laws in diffusion models has recently become a prominent focus within the research community~\citep{ma2025inference, yetfg,liu2024alignment}. This line of work can be traced back to Re-Sampling~\citep{lugmayr2022repaint}, which iteratively refines latent variables by injecting random Gaussian noise, effectively reverting the noise level to a previous scale. This iterative paradigm has been utilized in subsequent works, including universal conditional control~\citep{bansal2023universal}, video generation~\citep{wu2023tune}, and protein design~\citep{jumper2021highly}, to enhance inference performance. However, it has primarily been treated as a heuristic trick, with its underlying mechanisms remaining underexplored. Z-Sampling~\citep{bai2024zigzag} extended this paradigm by replacing random noise injection with inversion operations and identified the guidance difference between denoising and inversion as a critical factor. This phenomenon has also been validated in subsequent studies~\citep{zhou2024golden, shao2024iv, ahn2024noise}. In our work, we systematically unify these inference enhancement methods, demonstrating that their essence lies in approximating the strong-to-ideal difference via the weak-to-strong difference, and integrate them into a unified reflective framework, W2SD, through theoretical and empirical analysis. 

\section{Experiment Settings}
\label{sec:exp_setting}
In this section, we introduce the details of hyperparameters, metrics and datasets used in the experiments.
\subsection{Hyperparameters}
\paragraph{Weight Difference} 
For the W2SD based on weight difference, we consider full-parameter tuning, LoRA-based efficient tuning, and the MoE mechanism.

For the full-parameter tuning, we first set SD1.5~\citep{rombach2022high} as the weak model and the fine-tuned DreamShaper v8 as the strong model, which demonstrates superior performance in terms of human preference and achieves high-quality generation results. Similarly, we also use SDXL~\citep{podellsdxl} as the weak model and Juggernaut-XL as the strong model to further evaluate W2SD's performance under weight differences. 

For the efficient tuning, we distinguish strong and weak models by adjusting the LoRA scale. First, we use the xlMoreArtFullV1 LoRA checkpoint to test W2SD's ability to enhance overall image quality. Additionally, to validate the ability of W2SD to improve personalization, we select a series of personalized LoRAs, as detailed in~\cref{sec:exp_dataset}. For the strong model, the LoRA scale is set to 0.8, while for the weak model, it is set to -1.5.

Following the default settings~\citep{bai2024zigzag}, we set the denoising steps $T=50$, and the guidance scale to 5.5. Reflection operations steps $\lambda=T-1$. Notably, to eliminate influence from guidance differences, the guidance scale of $M^{\mathrm{s}}$ and $M^{\mathrm{w}}$ are both set to 1.0 during reflection, ensuring the guidance difference is zero. 

For the MoE (Mixture of Experts) mechanism, we select DiT-MoE-S~\citep{fei2024scaling} as the strong model, which routes the top 2 optimal experts out of 8 during the inference process. The weak model, in contrast, is configured to route the 2 least optimal experts out of 8. Following the default settings, the denoising steps $T=50$, and the guidance scale is set to 1.5.

\paragraph{Condition Difference} In the W2SD research based on condition differences, we focus on analyzing the differences caused by two mechanisms: guidance scale and prompt semantics.

By adjusting the guidance scale to distinguish the strong and weak model, we adapt the same settings as Z-Sampling~\citep{bai2024zigzag}: the guidance scale of $\mathcal{M}^{\mathrm{s}}$ was set to 5.5, and the guidance scale of $\mathcal{M}^{\mathrm{w}}$ is set to 0. The diffusion model used is SDXL.

For semantic differences, we set $\mathcal{M}^{\mathrm{w}}$ to use the GenEval prompt, which is short (4-5 words), ambiguous, and coarse, often resulting in uncontrolled outputs. In contrast, $\mathcal{M}^{\mathrm{s}}$ uses refined prompts enhanced by QWen-Plus~\citep{yang2024qwen2}, providing greater detail and semantic richness. During the reflection process, the guidance scale was set to 1.0 to eliminate the influence of guidance differences on the results.

Similar to the weight difference setup, we set $T=50$, $\lambda = T-1$, and the denoising guidance scale to 5.5.

\paragraph{Sampling Pipeline Difference} We demonstrate that W2SD can also generalize to capability differences across different diffusion pipelines. Specifically, we select ControlNet~\citep{zhang2023adding} as $\mathcal{M}^{\mathrm{s}}$, with the control scale set to the default value of 1.0. The standard sampling pipeline~\citep{song2020denoising} is chosen as the weak model. Consistent with the weight difference setup, we configure $T=50$, $\lambda = T-1$, and guidance scale of 5.5. During reflection operation, the guidance scale for both
$\mathcal{M}^{\mathrm{s}}$ and $\mathcal{M}^{\mathrm{w}}$ are set to 0.5 to eliminate the influence of guidance differences.

\subsection{Metrics}
\paragraph{AES.} AES~\citep{schuhmann2022laion} is an evaluation metric that assesses the visual quality of generated images by analyzing key aesthetic attributes such as contrast, composition, color, and detail, thereby measuring their alignment with human aesthetic standards.

\paragraph{PickScore.} PickScore~\citep{kirstain2023pick}  is a CLIP-based metric model trained on a comprehensive open dataset containing text-to-image prompts and corresponding real user preferences for generated images, specifically designed for predicting human aesthetic preferences.

\paragraph{HPS v2.} Building upon the Human Preference Dataset v2 (HPD v2), a comprehensive collection of 798,090 human preference judgments across 433,760 image pairs, \citep{wu2023human} developed HPS v2 (Human Preference Score v2) through CLIP fine-tuning, establishing a more accurate predictive model for human preferences in generated images.

\paragraph{MPS.} MPS~\citep{zhang2024learning} is a metric for text-to-image model evaluation, trained on the MHP dataset containing 918,315 human preference annotations across 607,541 images. This novel metric demonstrates superior performance by effectively capturing human judgments across four critical dimensions: aesthetic quality, semantic alignment, detail fidelity, and overall assessment.

\subsection{Datasets}
\label{sec:exp_dataset}
\paragraph{Pick-a-Pic.} The Pick-a-Pic dataset~\citep{kirstain2023pick}, collected through user interactions with a dedicated web application for text-to-image generation, systematically records each comparison with a prompt, two generated images, and a preference label (indicating either a preferred image or a tie when no significant preference exists). Following~\citet{bai2024zigzag}, we utilize the initial 100 prompts as a representative test set, which provides adequate coverage to assess model performance.

\paragraph{Drawbench.} 
DrawBench~\citep{saharia2022photorealistic} is a comprehensive evaluation benchmark for text-to-image models, featuring approximately 200 text prompts across 11 distinct categories that assess critical capabilities including color rendering, object counting, and text generation.

\paragraph{GenEval.}Geneval~\citep{ghosh2024geneval} is an object-focused framework that evaluates image composition through object co-occurrence, position, count, and color. Using 553 prompts, it achieves 83\% agreement with human judgments on image correctness.

\paragraph{VBench} VBench~\citep{huang2023vbench} is a comprehensive benchmark for video generation models, featuring a hierarchical evaluation framework across multiple quality dimensions. It supports both automatic and human assessment, with VBench++ extending to text-to-video and image-to-video tasks while incorporating trustworthiness evaluation.

\paragraph{Peronalization Dataset} To evaluate the performance gains of W2SD in personalized generation, we selected 20 LoRA checkpoints from the Civitai platform, covering a diverse range of categories, including persons (e.g. Anne Hathaway, Scarlett Johansson), animals (e.g., Scottish Fold cat, prehistoric dinosaur), styles (e.g., Disney style, parchment style), anime characters (e.g. Sun Wukong, Bulma) and objects (e.g. cars).

\section{Supplementary experimental results}
In this section, we present more quantitative and qualitative results of W2SD.
\label{sec:exp_res}

\subsection{Quantitative Results}
\label{sec:exp_res_quant}
\begin{table*}[t]
\caption{Different types of model differences lead to improvements effects in different directions.}
\label{tab:meta_appendix}
\vskip 0.15in
\begin{center}
\begin{small}
\begin{footnotesize}
\renewcommand{\arraystretch}{1.5}
\begin{tabular}{l|cccr}
\toprule
Model Difference & $\mathcal{M}^{\mathrm{s}}$ & $\mathcal{M}^{\mathrm{w}}$ & Results \\
\midrule
\multirow{3}{*}{Weight Difference} & Finetune Mechanism & SDXL/SD1.5 & \cref{tab:full_finetune,tab:full_finetune_drawbench} \\
& LoRA Mechanism & SDXL/SD1.5 & \cref{tab:ckpt_gap_personal,tab:lora_finetune_pick,tab:lora_finetune_drawbench} \\
& Strong Experts (MoE) & Weak experts (MoE) & \cref{tab:moe} \\
\midrule
\multirow{2}{*}{Condition Difference} & High CFG & Low CFG & \cref{tab:cfg_gap,tab:guidance_drawbench} \\
& Refined Prompt & Raw Prompt & \cref{tab:prompt_gap} \\
\midrule
\multirow{2}{*}{Sampling Pipeline Difference} & ControlNet & Standard Pipeline (DDIM) & ~\cref{fig:main_case_controlnet} \\ & IP-Adapter & Standard Pipeline (DDIM) & ~\cref{fig:ip-adapter}\\
\bottomrule
\end{tabular}
\end{footnotesize}
\end{small}
\end{center}
\vskip -0.1in
\end{table*}

\begin{table}[!h]
\centering
\begin{minipage}{0.48\textwidth}
    \centering
\caption{Quantitative results of W2SD based on a full parameter fine-tuning strategy. Our method generates results better aligned with human preferences. Datasets: Drawbench.}
\label{tab:full_finetune_drawbench}
\begin{center}
\begin{small}
\resizebox{1\textwidth}{!}{
\begin{tabular}{c|cccc}
\toprule
Method & HPS v2 $\uparrow$ & AES $\uparrow$ & PickScore $\uparrow$& MPS $\uparrow$\\
\midrule
SD1.5 & 25.3601 & 5.2023 & 21.0519 & - \\
\midrule
DreamShaper & 28.7845 & 5.7047 & 21.8522 & 47.8813  \\
\textbf{W2SD} & \textbf{28.7901} & \textbf{5.7847} & \textbf{21.9057} & \textbf{52.1192} \\
\midrule
\midrule
SDXL & 28.5536 & \textbf{5.4946} & 22.2464 & -  \\
\midrule
Juggernaut-XL & 28.9085 & 5.3455 & 22.4906 & 47.5648 \\
\textbf{W2SD}  & \textbf{29.3246} & 5.4261 & \textbf{22.5803} & \textbf{52.4358} \\
\bottomrule
\end{tabular}
}
\end{small}
\end{center}
\end{minipage}
\hfill
\begin{minipage}{0.48\textwidth}
\centering
\caption{Quantitative results of W2SD based on guidance difference. Model: SDXL. Datasets: DrawBench.}

\label{tab:guidance_drawbench}
\begin{center}
    \begin{small}
\resizebox{1\textwidth}{!}{
\begin{tabular}{c|cccc}
\toprule
Method & HPS v2 $\uparrow$& AES $\uparrow$& PickScore $\uparrow$& MPS $\uparrow$\\
\midrule
SD1.5 & 25.3601 & 5.2023 & 21.0519 & 47.7075 \\
\textbf{W2SD} & \textbf{25.8234} & \textbf{5.2157} & \textbf{21.2079} & \textbf{52.2934} \\
\midrule
SDXL & 28.5536 & \textbf{5.4946} & 22.2464 & 40.9590\\
\textbf{W2SD} & \textbf{30.1426} & \textbf{5.6600} & \textbf{22.4434} & \textbf{59.0415}\\
\bottomrule
\end{tabular}
}
    \end{small}
\end{center}
\end{minipage}
\end{table}

\begin{table}[!h]
\centering
\begin{minipage}{0.48\textwidth}
    \centering
\caption{Quantitative results of W2SD based on human preference LoRA model. Our method generates results better aligned with human preferences. Datasets: Pick-a-Pic.}
\label{tab:lora_finetune_pick}
\begin{center}
\begin{small}
\resizebox{1\textwidth}{!}{
\begin{tabular}{c|cccc}
\toprule
Method & HPS v2 $\uparrow$ & AES $\uparrow$ & PickScore $\uparrow$& MPS $\uparrow$\\
\midrule
SD1.5 & 24.9558 & 5.5003 & 20.1368& - \\
\midrule
Dpo-Lora & 25.5678 & 5.5804 & 20.3514 & 44.2889  \\
\textbf{W2SD} & \textbf{26.0825} & \textbf{5.6567} & \textbf{20.5096} & \textbf{55.7106} \\
\midrule
\midrule
SDXL & 29.8701 & 6.0939 & 21.6487 & -  \\
\midrule
xlMoreArtFullV1 & 32.8040 & 6.1176 & 22.3259 & 48.2224 \\
\textbf{W2SD}  & \textbf{33.5959} & \textbf{6.2252} & \textbf{22.3644} & \textbf{51.7770} \\
\bottomrule
\end{tabular}
}
\end{small}
\end{center}
\end{minipage}
\hfill
\begin{minipage}{0.48\textwidth}
\centering
\caption{Quantitative results of W2SD based on human preference LoRA model.  Our method generates results better aligned with human preferences. Datasets: DrawBench.}

\label{tab:lora_finetune_drawbench}
\begin{center}
    \begin{small}
\resizebox{1\textwidth}{!}{
\begin{tabular}{c|cccc}
\toprule
Method & HPS v2 $\uparrow$ & AES $\uparrow$ & PickScore $\uparrow$& MPS $\uparrow$\\
\midrule
SD1.5 & 25.3601 & 5.2023 & 21.0519 & - \\
\midrule
Dpo-Lora & 25.8896 & 5.2895 & 21.2308 & 49.3617  \\
\textbf{W2SD} & \textbf{25.9431} & \textbf{5.3553} & \textbf{21.2589} & \textbf{50.6399} \\
\midrule
\midrule
SDXL & 28.5536 & 5.4946 & 22.2464 & - \\
\midrule
xlMoreArtFullV1 & 31.2727 & 5.5487 & 22.7721 & 47.0396 \\
\textbf{W2SD}  & \textbf{32.34857} & \textbf{5.7595} & \textbf{22.8301} & \textbf{52.9588} \\
\bottomrule
\end{tabular}
}
    \end{small}
\end{center}
\end{minipage}
\end{table}

\paragraph{Results of W2SD in other benchmarks} To further validate the effectiveness of W2SD, we also conducted experiments on Drawbench~\citep{saharia2022photorealistic}. For clarity, we have systematically organized and summarized these experiments in~\cref{tab:meta_appendix}. In Drawbench, we report results based on weight difference (see~\cref{tab:full_finetune_drawbench,tab:lora_finetune_drawbench}) and guidance difference (see~\cref{tab:guidance_drawbench}). Additionally, in Pick-a-Pic, we report results based on weight difference (see~\cref{tab:lora_finetune_pick}). Notably, W2SD demonstrates consistent improvements across all evaluated metrics.

\paragraph{Results of W2SD in Video Generation Task}  We also validate the performance of W2SD on video generation task to demonstrate its broad applicability. We randomly select 200 prompts from VBench~\citep{huang2023vbench} as test prompt cases and focus on analyzing the AnimateDiff~\citep{guoanimatediff} as video generation model. 

For the strong model $\mathcal{M}^{\mathrm{s}}$,  we employ AnimateDiff with Juggernaut-XL using guidance scale of 3.0, while the weak model $\mathcal{M}^{\mathrm{w}}$ utilizes AnimateDiff with SDXL at guidance scale of 1.0. This setup introduces both weight and guidance differences, meeting the conditions required by W2SD.

\begin{table}[t]
\caption{Quantitative results of W2SD on the video generation task. Model: AnimateDiff. Datasets: VBench.}
\label{tab:vbench}
\begin{center}
\begin{small}
\resizebox{1.0\textwidth}{!}{
\begin{tabular}{c|cccccc}
\toprule
Method & Subject Consistency $\uparrow$& Background Consistency $\uparrow$& Motion Smoothness $\uparrow$ & Dynamic Degree $\uparrow$ & Aesthetic Quality $\uparrow$ & Image Quality $\uparrow$\\
\midrule
AnimateDiff (SDXL) & 91.4152\% & 95.8491\% & 94.1647\% & \textbf{45.0000\%} & 55.3108\% & 58.4293\%\\
AnimateDiff (Juggernaut-XL) & 96.0820\% & 97.5463\% & \textbf{96.8834\%} & 13.0653\% & 57.7097\% & 64.1674\%\\
\midrule
\textbf{W2SD} & \textbf{97.1398\%} & \textbf{97.9386\%} & 96.8706\% & 8.0000\% & \textbf{59.3736\%} & \textbf{64.6987\%}\\
\bottomrule
\end{tabular}
}
\end{small}
\end{center}
\end{table}

In~\cref{tab:vbench}, W2SD achieves significant improvements across different dimensions such as Subject Consistency, Background Consistency, Aesthetic Quality and mage Quality, confirming its effectiveness in video generation task. Notably, in the Motion Smoothness dimension, W2SD exhibits a performance degradation due to Juggernaut-XL's inherent limitations compared to SDXL. This observation further validates our theory in~\cref{sec: theory}.

\subsection{Qualitative Results}
\label{sec:exp_res_quali}
\begin{figure}
    \centering
    \includegraphics[width=0.95\linewidth]{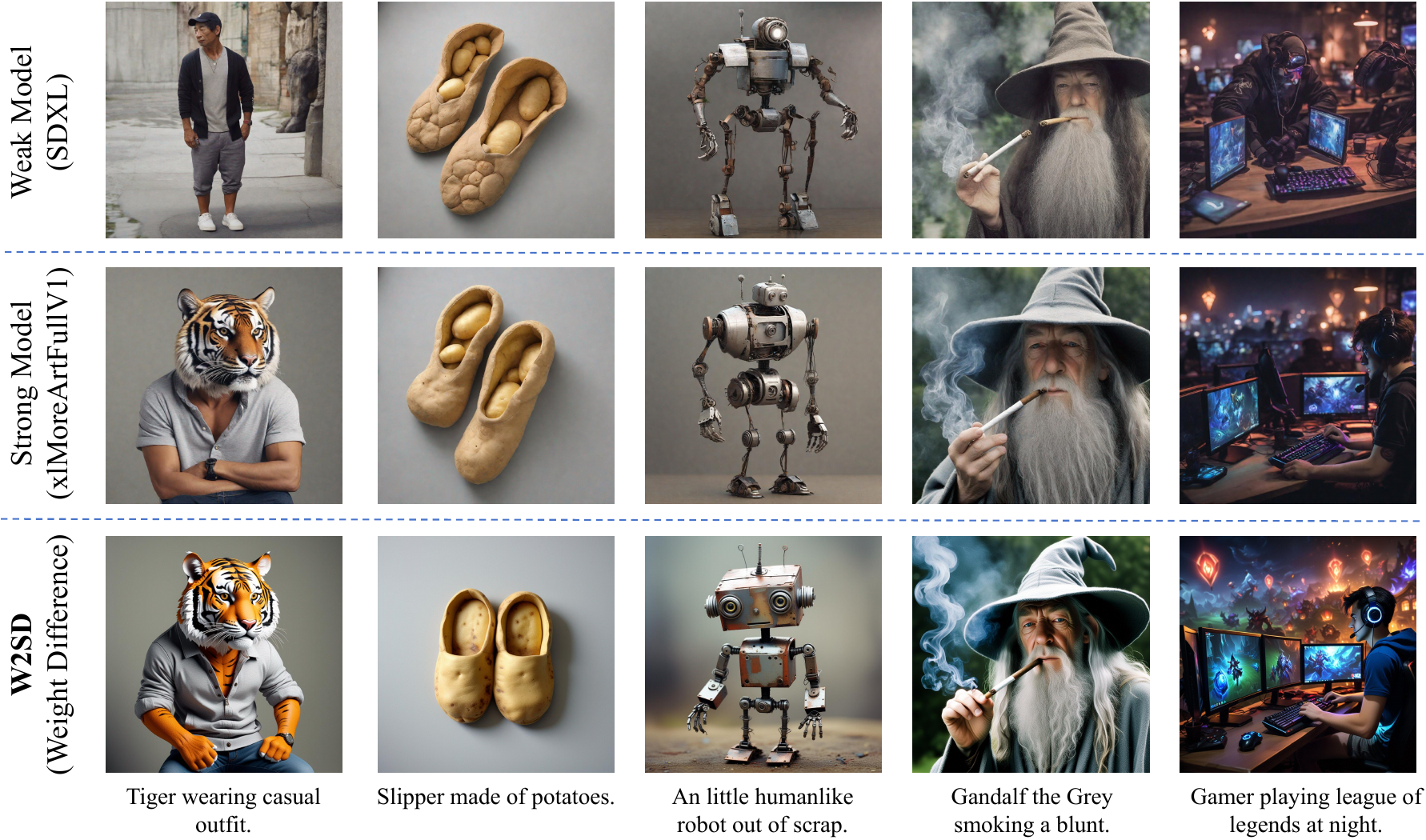}
    \caption{Qualitative results of W2SD based on weight differences (human preference). Here we select xlMoreArtFullV1 as the strong model and SDXL as the weak model. W2SD can effectively enhance the performance of human preference.}
    \label{fig:weight_preference_appendix}
\end{figure}

\paragraph{Weight Difference} In~\cref{fig:weight_preference_appendix}, we present visualization results of W2SD based on weight difference. Specifically, to enhance human preference for generated images, we select xlMoreArtFullV1 as the strong model and SDXL as the weak model.

\begin{figure}
    \centering
    \includegraphics[width=0.95\linewidth]{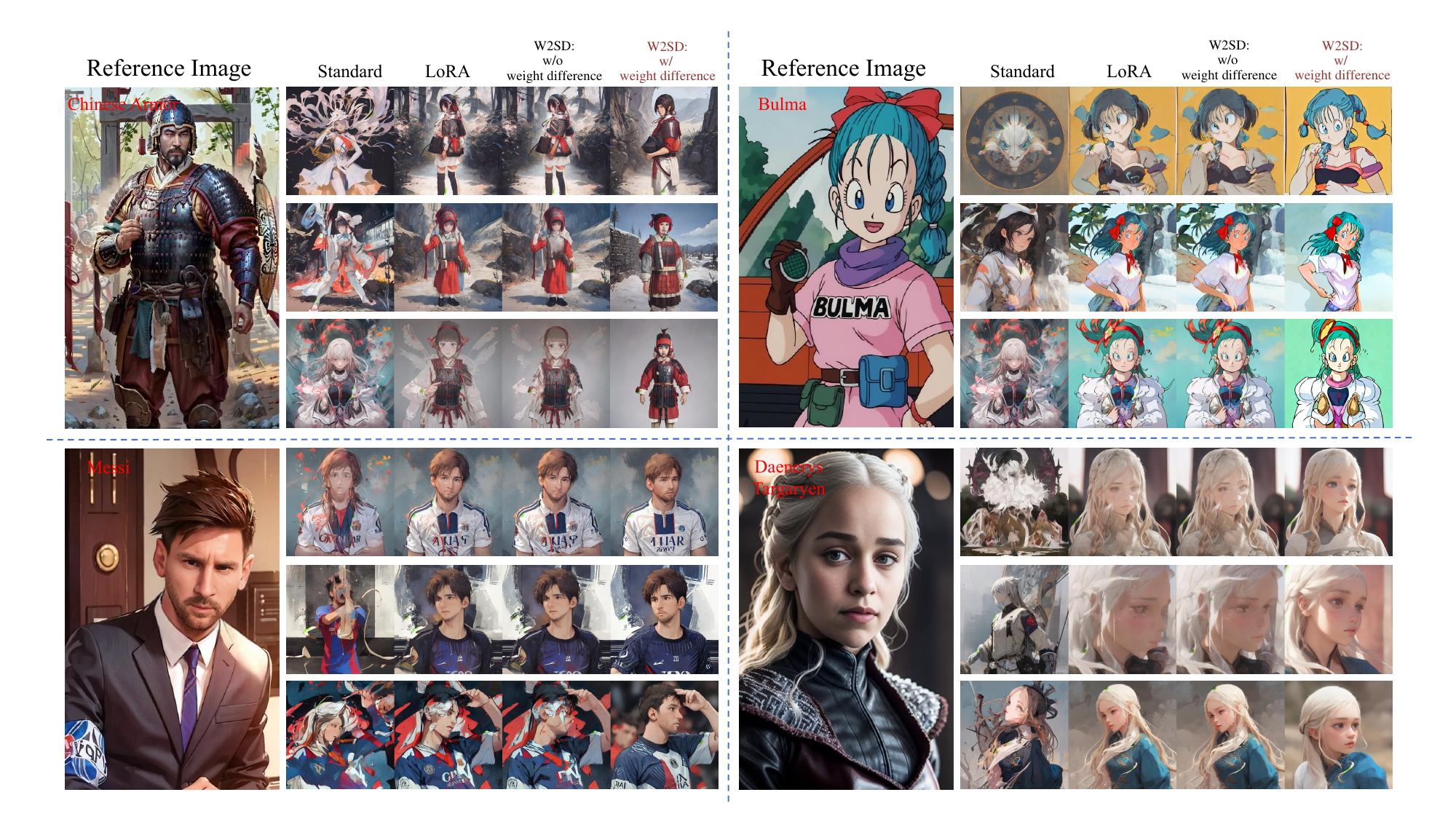}
    \caption{Qualitative results of W2SD based on weight differences (personalization). Here we set LoRA-based personalized model as strong model and the standard model as weak model}
    \label{fig:weight_personalization_appendix}
\end{figure}

Additionally, by setting the LoRA-based personalized model as strong model and the un-finetuned base model as weak model,~\cref{fig:weight_personalization_appendix} showcases the improvement of W2SD in personalized generation effectiveness.

\begin{figure}[h]
    \centering
    \includegraphics[width=1\linewidth]{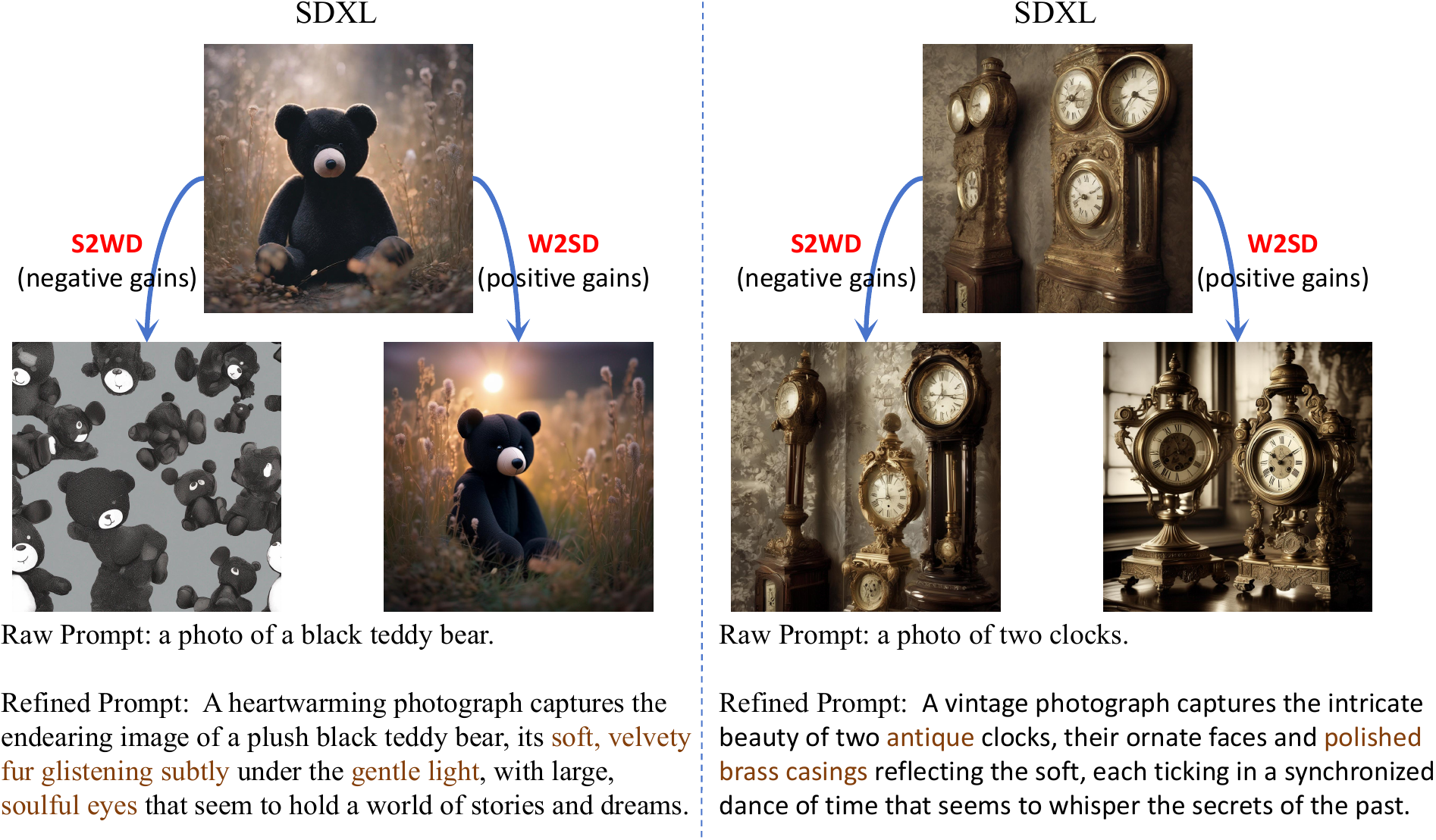}
    \caption{Qualitative results of W2SD based on semantic differences between prompts, which refines the generation process by placing greater emphasis on the fine-grained details.}
    \label{fig:case_prompt_gap}
\end{figure}

\paragraph{Condition Difference} In~\cref{fig:case_prompt_gap}, we present the visualization results of W2SD based on condition differences. When the strong model utilizes detailed and semantically rich prompts, while the weak model relies on simple prompts (containing only 4–5 words), W2SD effectively captures fine-grained conditional features. 
Notably, Z-Sampling~\citep{bai2024zigzag} is a special case of W2SD based on guidance differences, with extensive visual evidence already provided; thus, we omit additional visualizations here.

\begin{figure}[h]
    \centering
    \includegraphics[width=1\linewidth]{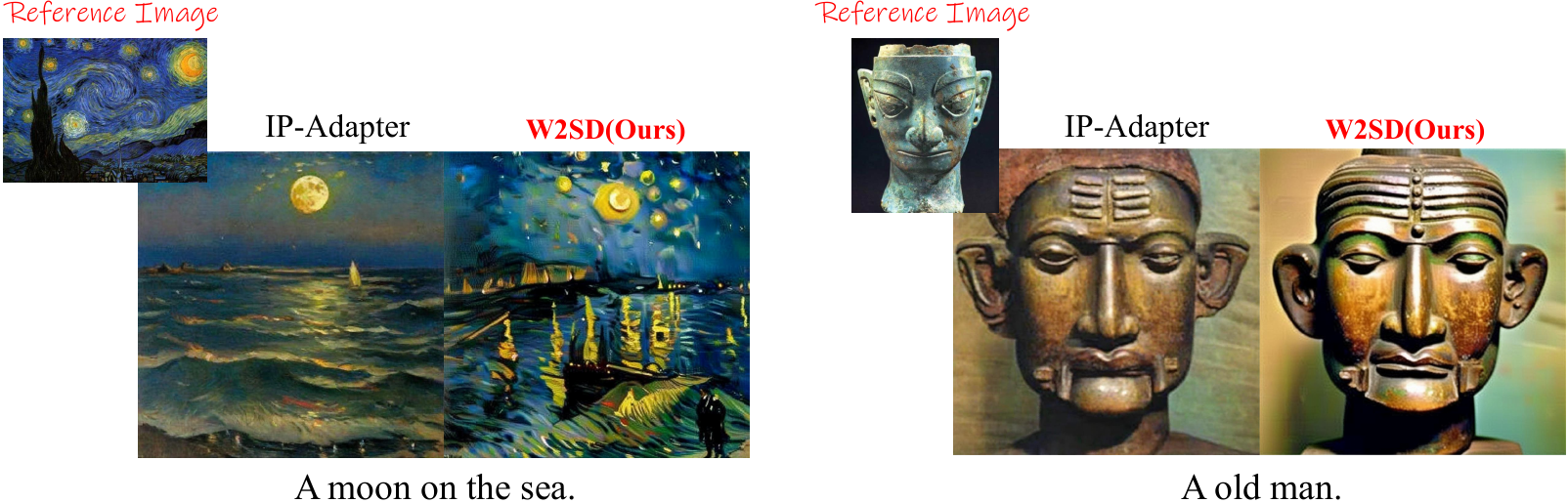}
    \caption{Qualitative results of W2SD based on sampling pipeline difference. When the strong model employs Ip-adapter and the weak model utilizes DDIM , W2SD can enhance the alignment of the generated results with the reference image (e.g., style).}
    \label{fig:ip-adapter}
\end{figure}

\paragraph{Sampling Pipeline Difference} In~\cref{fig:ip-adapter}, we present the visualization results of W2SD based on the differences in the sampling pipeline. By incorporating reference image information through Ip-adapter during the denoising process and employing standard DDIM for inversion, W2SD ensures that the generated image adheres more closely to the given stylistic conditions, resulting in higher-quality results.
\clearpage
\section{Proofs}
\label{sec:proof}
In this section, we provide the proofs for~\cref{theorem:1} presented in this work.

\subsection{Proof of Theorem~\ref{theorem:1}}
\label{proof:1}
We first establish the relationship between the latent variable $x_{t}$ and the refined latent variable $\Tilde{x}_{t}$ through the reflection operation, proving that the core mechanism of W2SD is to optimize $x_{t}$
in the direction of the weak-to-strong difference.
\begin{proof}
At a given time $t$, the W2SD reflection operation applies the operator $\mathcal{M}^{w}_{inv}\mathcal{M}^{s}(\cdot)$ to the latent variable $x_{t}$.

Specifically, we first denoise $x_{t}$ using the strong model $\mathcal{M}^{\mathrm{s}}$ according to~\cref{eq:backward_ode}, obtaining $x_{t-\Delta t}$ as 
\begin{equation}
    x_{t-\Delta t}
    = x_{t} + \sigma^{2t}s_{\theta}^{\mathrm{s}}(x_{t},t)\Delta t,
    \label{eq:Ms_1_appendix}
\end{equation}
where $s_{\theta}^{\mathrm{s}}$ denotes the score predicted by $\mathcal{M}^{\mathrm{s}}$, which is equivalent to the gradient of the log probability density $\log{p_{t}^{\mathrm{s}}}$, so,~\cref{eq:Ms_1_appendix} can be rewritten as
\begin{equation}
    x_{t-\Delta t}
    = x_{t} + \sigma^{2t}\nabla_{x_{t}} \log{p_{t}^{\mathrm{s}}(x_{t})}\Delta t.
    \label{eq:Ms_2_appendix}
\end{equation}

After obtaining $x_{t-\Delta t}$, we apply the weak model $\mathcal{M}^{\mathrm{w}}$ to invert $x_{t-\Delta t}$ back to the noise level at time 
$t$ according~\cref{eq:inversion_ode_approx}, thereby completing the reflection process as

\begin{equation}
    \Tilde{x}_{t} \approx x_{t-\Delta t} - \sigma^{2t}s_{\theta}^{\mathrm{w}}(x_{t-\Delta t}, t)\Delta t,
    \label{eq:Mw_1_appendix}
\end{equation}
where $s_{\theta}^{\mathrm{w}}$ denotes the score predicted by $\mathcal{M}^{\mathrm{w}}$. Since $\Delta t$ is typically small, we neglect the approximation error in the diffusion inversion process which implies that $s_{\theta}^{\mathrm{w}}(x_{t-\Delta t},t)$ is equivalent to the gradient of the log probability density $\log{p_{t}^{\mathrm{w}}}$. Hence we can reformulate~\cref{eq:Mw_1_appendix} as 
\begin{equation}
    \Tilde{x}_{t} = x_{t-\Delta t} - \sigma^{2t}\nabla_{x_{t}} \log{p_{t}^{\mathrm{w}}(x_{t})}\Delta t.
    \label{eq:Mw_2_appendix}
\end{equation}

Combining~\cref{eq:Ms_2_appendix} and~\cref{eq:Mw_2_appendix}, we obtain $\Tilde{x}_{t}$ as
\begin{align}
  \Tilde{x}_{t} &= x_{t-\Delta t} - \sigma^{2t}\nabla_{x_{t}} \log{p_{t}^{\mathrm{w}}(x_{t})}\Delta t\\ 
  &=  x_{t} + \sigma^{2t}\nabla_{x_{t}} \log{p_{t}^{s}(x_{t})}\Delta t - \sigma^{2t}\nabla_{x_{t}} \log{p_{t}^{\mathrm{w}}(x_{t})}\Delta t\\
  &= x_{t} + \sigma^{2t}\Delta t (\underbrace{\nabla_{x_{t}} \log{p_{t}^{s}(x_{t})} - \nabla_{x_{t}} \log{p_{t}^{\mathrm{w}} (x_{t})}}_{\text{weak-to-strong difference} \Delta_{1} (t)} ).
  \label{eq:ref_effect}
\end{align}

From~\cref{eq:ref_effect}, we observe that for the latent variable $x_{t}$, the reflection operator $\mathcal{M}^{\mathrm{w}}_{\mathrm{inv}}(\mathcal{M}^{\mathrm{s}}(\cdot))$ in W2SD perturbs $x_{t}$ along the direction of $\Delta_{1}(t)$, producing the refined variable $\Tilde{x}_{t}$. When the weak-to-strong difference $\Delta_{1}$ closely bridges the unattainable strong-to-ideal difference $\Delta_{2}(t)$, the resulting $\Tilde{x}_{t}$ becomes more aligned with the ground truth ideal distribution.
\end{proof}

\section{Further analytical exploration}
\label{sec:further_analys}

\subsection{Deeper Visual Understanding}
\label{sec:deeper_visual_understanding}
\begin{figure}
    \centering
    \includegraphics[width=0.75\linewidth]{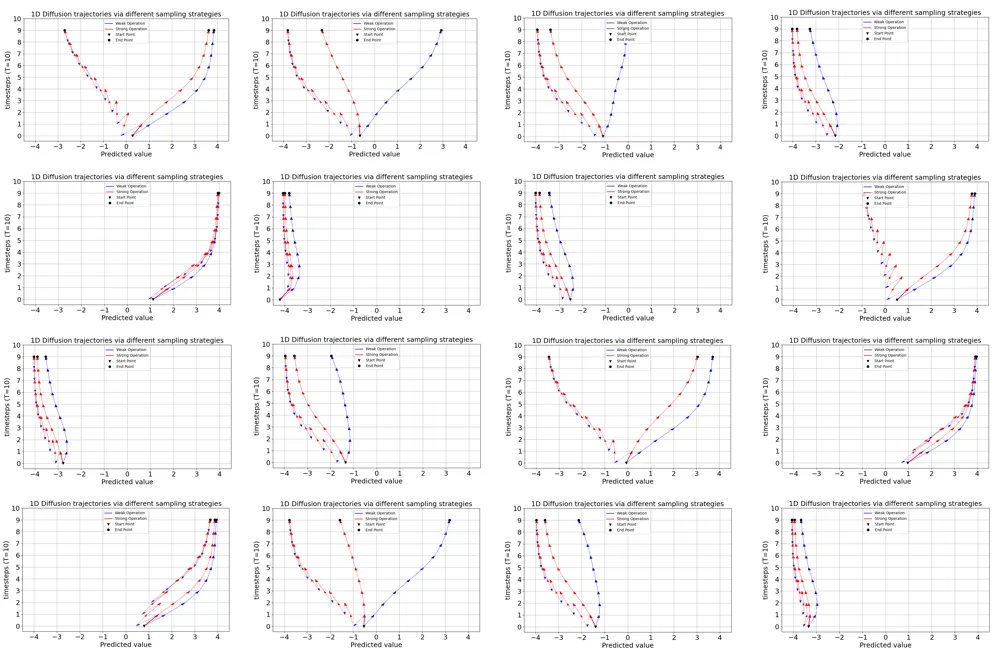}
    \caption{Denoising trajectories across different settings (1-D Gauss). The weak model (blue) generates only right-peak data due to missing left-peak training samples, while the strong model (red) produces data between both peaks. W2SD balances the distribution by leveraging the reflective operator $\mathcal{M}_{\mathrm{inv}}^{\mathrm{w}}(\mathcal{M}^{\mathrm{s}}(\cdot))$.}
    \label{fig:1d-more-case}
\end{figure}

To provide a more intuitive demonstration of the W2SD effect, we present more 1D visualization examples in~\cref{fig:1d-more-case} to aid understanding. As can be seen, W2SD has a certain probability of ``correcting'' the denoising trajectory of the model, resulting in more balanced generation outcomes. The settings here are consistent with those in~\cref{fig:1d-gauss}.

We also empirically validate that the weak-to-strong difference $\Delta_{2}$ can effectively bridges the strong-to-ideal difference $\Delta_{1}$. We first compute $\mathbf{CosineSimilarity}(\Delta_{1}(t),\Delta_{2}(t))$ at each timestep $t$. We train three models on CIFAR-10: the weak model (100 epochs), the strong model (300 epochs), and the ideal model (fully converged, 500 epochs). Here $\Delta_{1}(t) = \epsilon_{ideal}(t) - \epsilon_{strong}(t)$ and $\Delta_{1}(t) = \epsilon_{strong}(t) - \epsilon_{weak}(t)$. In the table below, the angle between $\Delta_{1}(t)$ and $\Delta_{2}(t)$ remains below $90^{o}$, confirming that weak-to-strong gap can reliably bridge strong-to-ideal gap.

\begin{figure}[h]
    \centering
    \includegraphics[width=0.5\linewidth]{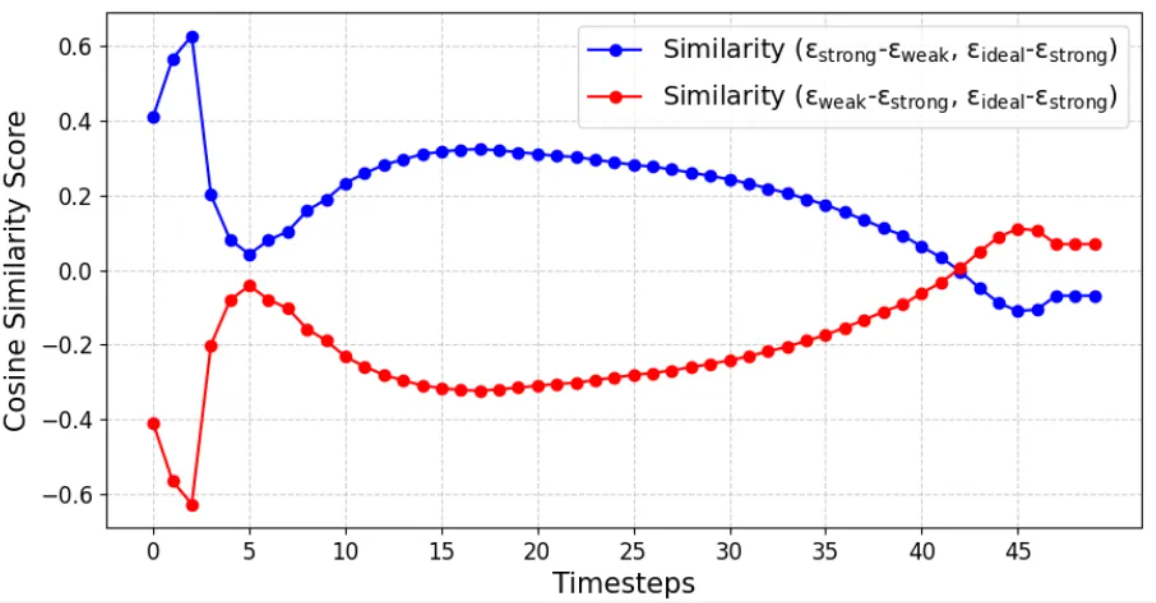}
    \caption{Positive cosine similarity between $\Delta_{1}$ and $\Delta_{2}$ ($\cos{\Theta}<90$°).}
    \label{fig:enter-label}
\end{figure}

\begin{figure}[h]
    \centering
    \includegraphics[width=1\linewidth]{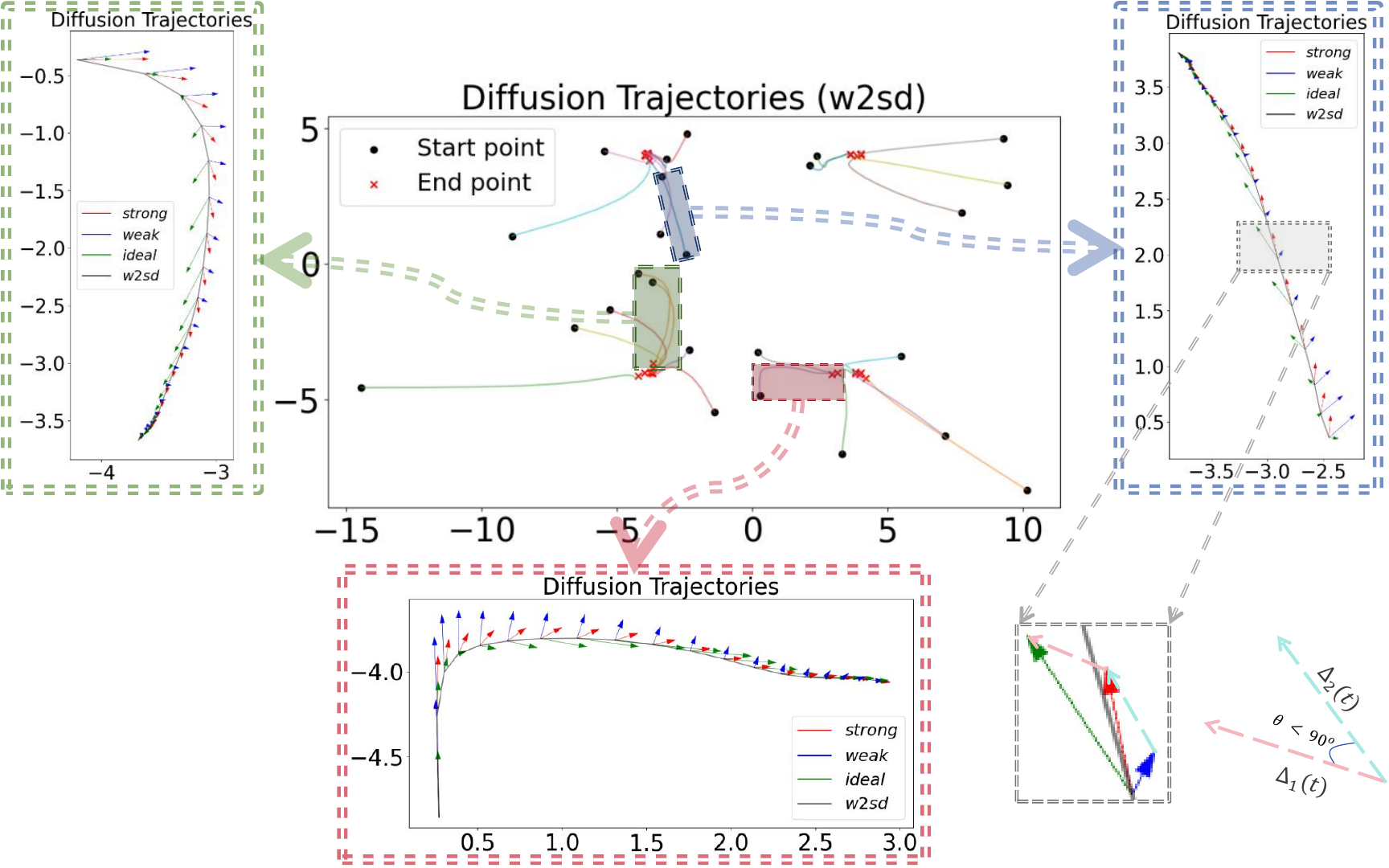}
    \caption{2D denoising path visualization. The weak-to-strong gap can effectively bridge the strong-to-ideal gap.}
    \label{fig:more_visual_case_path}
\end{figure}

To further validate this claim that $\Delta_{2}$ can bridge $\Delta_{1}$, we provide additional visualizations. As illustrated in~\cref{fig:more_visual_case_path}, the W2SD trajectory is visualized in 2D space, where its denoising direction (black) demonstrates closer alignment with the ideal direction (green). Moreover, the angles between the strong (red)-to-weak (blue) and ideal (green)-to-strong (red) vectors consistently remain within 90 °.

\subsection{The Impact of Approximation Error in the Inversion Process}
\label{sec:error_exp}

\begin{table}[!h]
\centering
\begin{minipage}{0.48\textwidth}
    \centering
\caption{As the magnitude of the approximation error in inversion process increases, the gains from W2SD diminish. Model :xlMoreArtFullV1. Datasets: Pick-a-Pic. The type of W2SD: weight difference.}
\label{tab:inversion_error}
\begin{center}
\begin{small}
\resizebox{1\textwidth}{!}{
\begin{tabular}{c|cccc}
\toprule
Method & HPS v2 $\uparrow$ & AES $\uparrow$ & PickScore $\uparrow$& MPS $\uparrow$\\
\midrule
SDXL & 29.8701 & 6.0939 & 21.6487 & -  \\
xlMoreArtFullV1 & 32.8040 & 6.1176 & 22.3259 & 48.2224\\
\midrule
\textbf{W2SD (k=0)} & \textbf{33.5959} & 6.2252 & \textbf{22.3644} & \textbf{51.7770}\\
W2SD (k=0.005) & 32.9341 & \textbf{6.3228} & 22.3221 & 46.7130   \\
W2SD (k=0.010)  & 19.5277 & 4.3949 & 17.9267 & 1.3440  \\
\bottomrule
\end{tabular}
}
\end{small}
\end{center}
\end{minipage}
\hfill
\begin{minipage}{0.48\textwidth}
\centering
\caption{By selecting Gaussian noise during the Re-Sampling process, the advanced~\cref{algo:advanced_resampling} achieves superior performance in the sampling process, demonstrating that Re-Sampling is a specific instance of W2SD. Model: SDXL. Datasets: Pick-a-Pic. The type of W2SD: guidance difference.}

\label{tab:resampling_res}
\begin{center}
    \begin{small}
\resizebox{1\textwidth}{!}{
\begin{tabular}{c|cccc}
\toprule
Method & HPS v2 $\uparrow$ & AES $\uparrow$ & PickScore $\uparrow$& MPS $\uparrow$\\
\midrule
SDXL & 29.8701 & 6.0939 & 21.6487 & -  \\
\midrule
Re-Sampling (sim score$<$0) & 30.2797 & 6.0744 & 21.6894 & 48.4793  \\
Re-Sampling (sim score$>$0) & 30.7844 & 6.0555 & \textbf{21.8620} & 51.5210  \\
\midrule
\textbf{W2SD} & \textbf{31.2020} & \textbf{6.0970} & 21.7980 & \textbf{56.0608}  \\
\bottomrule
\end{tabular}
}
    \end{small}
\end{center}
\end{minipage}
\end{table}

In~\cref{sec:Preliminaries}, we assume that the inversion and denoising process are reversible, meaning that for the same generative model (including the same guidance scale, etc.), $\mathcal{M}_{\mathrm{inv}}(\mathcal{M}(x_{t},t),t)=x_{t}$. However, in practice, the inversion process inevitably introduces errors. Specifically, for $x_{t}$ at time $t$, the inversion process actually used in the algorithm implementation is as
\begin{align}
  \Tilde{x}_{t} &= x_{t-\Delta t} - \sigma^{2t}s_{\theta}(x_{t-\Delta t}, t)\Delta t
  \\
  &= x_{t-\Delta t} - \sigma^{2t}(\underbrace{s_{\theta}(x_{t-\Delta t}, t) - s_{\theta}(x_{t}, t)}_{\textrm{Inversion Error $\mathrm{E}_{t}$}} + s_{\theta}(x_{t}, t))\Delta t,
  \label{eq:abla_inv}
\end{align}
And the effect of W2S on the latent variable in~\cref{theorem:1} can also be expressed as
\begin{equation}
    \Tilde{x}_{t} = x_{t} + \sigma^{2t}\Delta t ( \nabla_{x_{t}} \log{p_{t}^{\mathrm{s}}(x_{t})} - \nabla_{x_{t}} \log{p_{t}^{\mathrm{w}} (x_{t})} - \textrm{E}_{t}),
\end{equation}
where $\textrm{Inversion Error $\mathrm{E}_{t}$}$ represents the approximation error in the inversion process at time $t$. Since $\Delta t$ is very small, $s_{\theta}(x_{t}, t)$ and $s_{\theta}(x_{t-\Delta t}, t)$ are assumed to be approximately equal by default, i.e., $\mathrm{E}_{t} \approx 0$.

To further analyze the impact of $\mathrm{E}_{t}$ on the W2SD, we simplify the analysis by replacing $\mathrm{E}_{t}$ with Gaussian noise of controllable magnitude as
\begin{equation}
    \Tilde{x}_{t} = x_{t} + \sigma^{2t}\Delta t ( \nabla_{x_{t}} \log{p_{t}^{\mathrm{s}}(x_{t})} - \nabla_{x_{t}} \log{p_{t}^{\mathrm{w}} (x_{t})} - k\epsilon).
\end{equation}
By varying the value of $k$, we adjust the magnitude of the error $\textrm{E}_{t}$ to study its effect on the generated results. We analyze W2SD based on weight differences, with the strong model using xlMoreArt-FullV1 and the weak model using SDXL. 

In~\cref{tab:inversion_error}, as $k$ increases, indicating a larger approximation error $\textrm{E}_{t}$, the gains from W2SD diminish. This finding is consistent with~\citet{bai2024zigzag}'s results, demonstrating that this phenomenon occurs in both guidance difference and weight difference scenarios, indicating that minimizing the approximation error in inversion process is crucial.

\begin{figure}[h]
\centering
\begin{minipage}{0.48\textwidth}
    \centering
    \includegraphics[width=\textwidth]{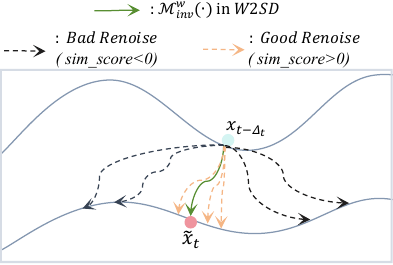}
    \caption{Re-Sampling, which can be considered a specific instance of W2SD, demonstrates improved performance when the randomly sampled Gaussian noise aligns closely with the perturbation vector introduced by the W2SD reflection mechanism (e.g., cosine similarity $>$ 0)}
    \label{fig:Resampling_framework}
\end{minipage}
\hfill 
\begin{minipage}{0.48\textwidth} 
    \centering
    \includegraphics[width=\textwidth]{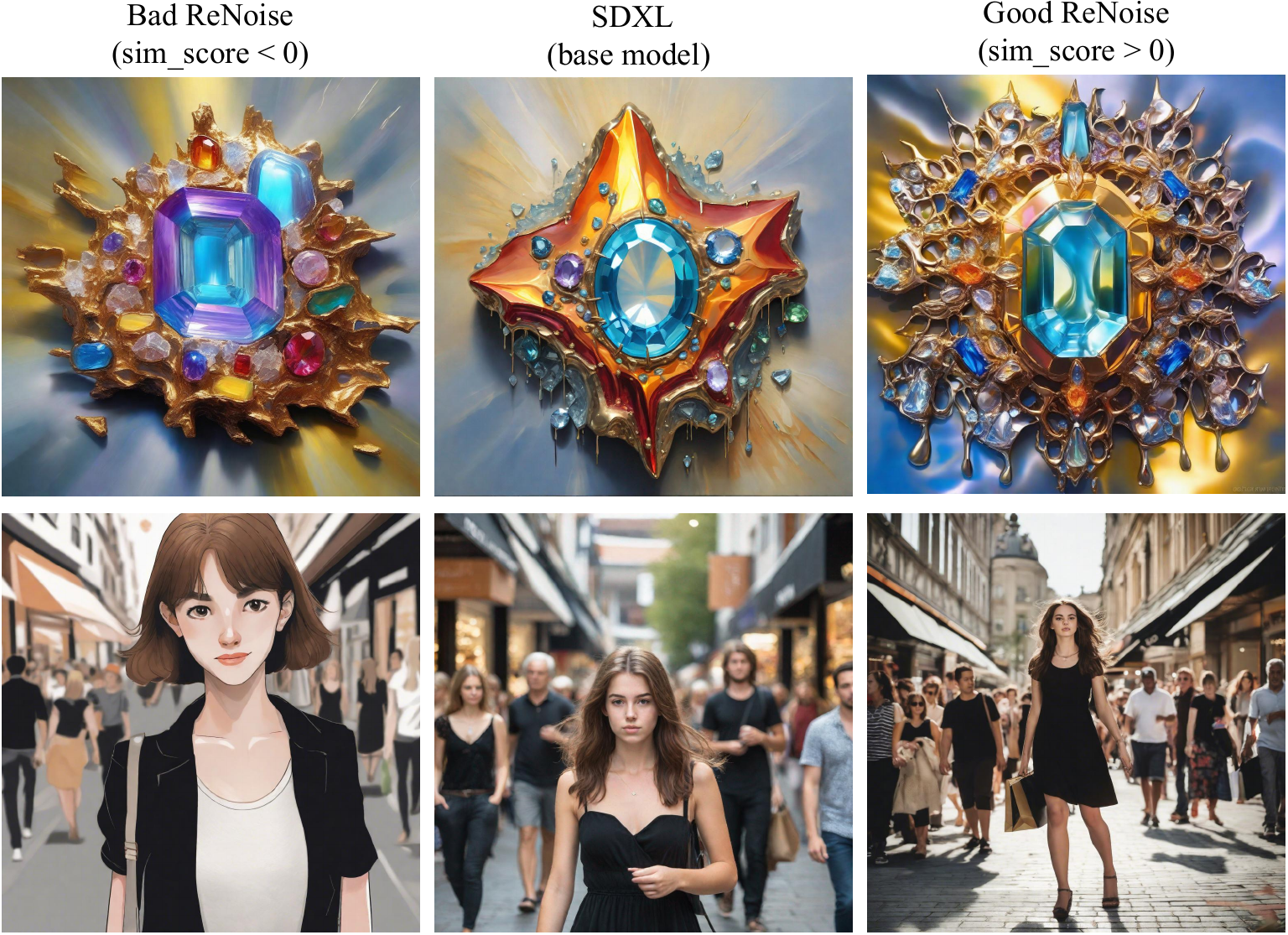}
    \caption{Qualitative results of advanced Re-Sampling demonstrate that the improvements effects vary depending on the strategy used to select Gaussian noise for Re-Sampling. It can be considered a specific instance of W2SD.}
    \label{fig:Resampling_case}
\end{minipage}
\end{figure}

\subsection{Relationship with Re-Sampling}
\label{sec:other_methods}
Re-Sampling~\citep{lugmayr2022repaint} is the earliest and most classic iterative algorithm of its kind proposed in diffusion models. As illustrated in~\cref{algo:vanilla_resampling}, it introduces randomly sampled Gaussian noise into the latent variable $x_{t}$, followed by repeated denoising processes.

We confirm that Re-Sampling is generally a specific instance of W2SD, which can be interpreted within the framework of our theory in~\cref{theorem:1}. In~\cref{fig:Resampling_framework}, we note that Re-Sampling performs better when the randomly sampled Gaussian noise in Re-Sampling is similar to the perturbation vector introduced by the W2SD reflection mechanism (e.g., cosine similarity $>$ 0). Additionally, we propose an advanced version of Re-Sampling (see~\cref{algo:advanced_resampling}), which evaluates whether the similarity score, \textbf{sim\_score}($\epsilon_{\mathrm{w2s}},\epsilon$), exceeds 0 to determine whether the Gaussian noise $\epsilon$ should be accepted for the ReNoise operation.

We validate the effectiveness of~\cref{algo:advanced_resampling} in~\cref{tab:resampling_res}. When consistently selecting favorable random noise (i.e., $\epsilon_{\mathrm{w2s}},\epsilon$) $>$ 0), advanced Re-Sampling demonstrates improved performance. Conversely, when consistently selecting unfavorable random noise (i.e., \textbf{sim\_score}($\epsilon_{\mathrm{w2s}},\epsilon$) $<$ 0), the performance of Re-Sampling deteriorates. 
We also present the visualization results in~\cref{fig:Resampling_case}, further demonstrating that Re-Sampling can be incorporated into the W2SD framework, validating the correctness of our theory in~\cref{theorem:1}.

Similarly, many subsequent research, such as FreeDoM~\citep{yu2023freedom}, UGD~\citep{bansal2023universal}, MPGD~\citep{he2023manifold}, and TFG~\citep{ye2024tfg}, follow the same approach as Re-Sampling by utilizing random Gaussian noise for iterative optimization. Therefore, these inference enhancement methods can be regarded as specific instances of W2SD. The primary distinction among these methods lies in the specific strong model $\mathcal{M}^{\mathrm{s}}$ they employ. For instance, TFG utilizes a more refined parameter search mechanism, resulting in a strong model that exhibits greater robustness and performance compared to algorithms such as FreeDoM and UGD. As a consequence, under the condition of the same weak model (i.e., the addition of random Gaussian noise), TFG demonstrates significantly enhanced performance. However, these studies collectively overlook a fundamental aspect: the weak-to-strong difference constitutes the core principle that fundamentally drives the efficacy of such algorithms.

\begin{figure}[h]
\centering
\begin{minipage}{0.48\textwidth}
    \centering
    \begin{algorithm}[H]
        \caption{Vanilla Re-Sampling}
        \begin{algorithmic}
        \label{algo:vanilla_resampling}
            \STATE {\bfseries Input:} Strong Model $\mathcal{M}^{\mathrm{s}}$, Total Inference Steps: $T$, Optimization Steps: $\lambda$
            \STATE {\bfseries Output:} Clean Data $x_{0}$
            \STATE Sample Gaussian noise $x_{T}$
            \FOR{$t=T$ {\bfseries to} $1$}
                \IF{$t > T - \lambda$}

                \STATE$x_{t-1} = \mathcal{M}^{\mathrm{s}}(x_{t},t)$
                
                \STATE Initialize $\epsilon \sim \mathcal{N}(0, 1)$
                \STATE $x^{\mathrm{Re}}_{t} = \texttt{Add\_Noise}(x_{t}, \epsilon, t)$
                \ENDIF
                \STATE $x_{t-1} = \mathcal{M}^{\mathrm{s}}(x^{\mathrm{Re}}_{t}, t)$
            \ENDFOR
        \end{algorithmic}
    \end{algorithm}
\end{minipage}
\hfill
\begin{minipage}{0.48\textwidth}
    \centering
    \begin{algorithm}[H]
        \caption{Advanced Re-Sampling}
        \begin{algorithmic}
        \label{algo:advanced_resampling}
            \STATE {\bfseries Input:} Strong Model $\mathcal{M}^{\mathrm{s}}$, Weak Model $\mathcal{M}^{\mathrm{w}}$, Total Inference Steps: $T$, Optimization Steps: $\lambda$
            \STATE {\bfseries Output:} Clean Data $x_{0}$
            \STATE Sample Gaussian noise $x_{T}$
            \FOR{$t=T$ {\bfseries to} $1$}
                \IF{$t > T - \lambda$}

                \STATE $x_{t-1} = \mathcal{M}^{\mathrm{s}}(x_{t},t)$

                \STATE \textcolor{red}{$\Tilde{x}_{t} = \mathcal{M}^{\mathrm{w}}_{\mathrm{inv}}(x_{t-1},t)$}
                \STATE \textcolor{red}{$\epsilon_{\mathrm{w2s}} = \Tilde{x}_{t} - x_{t}$}
                \STATE \textcolor{red}{Calculate $\epsilon_{\mathrm{w2s}}$ based on~\cref{eq:theorem_1_eq}}
                \STATE \#Select Optimal ReNoise
                \STATE Initialize $\epsilon         \sim \mathcal{N}(0, 1)$
                    \textcolor{red}{\WHILE{similarity\_score $(\epsilon_{\mathrm{w2s}},\epsilon)<0$}
                        \STATE Initialize $\epsilon         \sim \mathcal{N}(0, 1)$
                     \ENDWHILE}
                \STATE $x^{\mathrm{Re}}_{t} = \texttt{Add\_Noise}(x_{t}, \epsilon, t)$
                \ENDIF
                \STATE $x_{t-1} = \mathcal{M}^{\mathrm{s}}(x^{\mathrm{Re}}_{t}, t)$
            \ENDFOR
        \end{algorithmic}
    \end{algorithm}
\end{minipage}
\end{figure}

\subsection{Relationship with Auto-guidance}
\label{sec:auto-guidance}
We note that in the weak-to-strong framework, Auto-guidance~\citep{karras2024guiding} employs a pre-trained diffusion model along with a corrupted version of it (typically achieved by adding perturbations or reducing training iterations through training from scratch). It directly enhances performance by interpolating in the latent space. And here we clarify the contributions of W2SD in relation to it.

\paragraph{Different mechanisms} W2SD employs an reflection mechanism, while Auto-guidance utilizes an interpolation-based method. For comparison with W2SD, we set $w$=1 in Auto-guidance. When using Strong Model (human preference model) vs Weak Model (SDXL), in~\cref{tab:auto-guidance}, we show that W2SD achieves notably higher scores on human preference metrics including PickScore and HPS v2. However, in this setting, direct interpolation in Auto-guidance leads to performance degradation in certain metrics, manifesting as oversaturation and artifacts. Actually, the auto-guidance mechanism similarly utilizes the following operations as
\begin{align}
  &x_{t} \rightarrow x_{t-1}^{good}\\ 
  &x_{t} \rightarrow x_{t-1}^{bad}\\
  &x_{t-1}^{new} = x_{t-1}^{good} + w \cdot (x_{t-1}^{good}-x_{t-1}^{bad}).
  \label{eq:auto-guidance}
\end{align}
When $w$ is too large, applying~\cref{eq:auto-guidance} roughly refines the distribution of the latent variables, leading to an unnatural shift in the data distribution~\citep{sadat2024eliminating,lou2023reflected}. When $w$ is too small, it fails to produce sufficient refinement.

\begin{table}[!h]
\centering
\begin{minipage}{0.55\textwidth}
    \centering
\caption{Performance comparison between Auto-guidance's interpolation and W2SD's reflection mechanism in latent variable refinement.}
\label{tab:auto-guidance}
\begin{center}
\begin{small}
\resizebox{1\textwidth}{!}{
\begin{tabular}{c|cccc}
\toprule
Method & HPS v2 $\uparrow$ & AES $\uparrow$ & PickScore $\uparrow$\\
\midrule
SDXL & 29.8701 & 6.0939 & 21.6487   \\
xlMoreArtFullV1 & 32.8040 & 6.1176 & 22.3259\\
Auto-guidance & 32.1650 & 6.1187 & 22.0177\\
\midrule
\textbf{W2SD} & \textbf{33.5959} & \textbf{6.2252} & \textbf{22.3644}\\
\bottomrule
\end{tabular}
}
\end{small}
\end{center}
\end{minipage}
\end{table}

\begin{figure}
    \centering
    \includegraphics[width=0.8\linewidth]{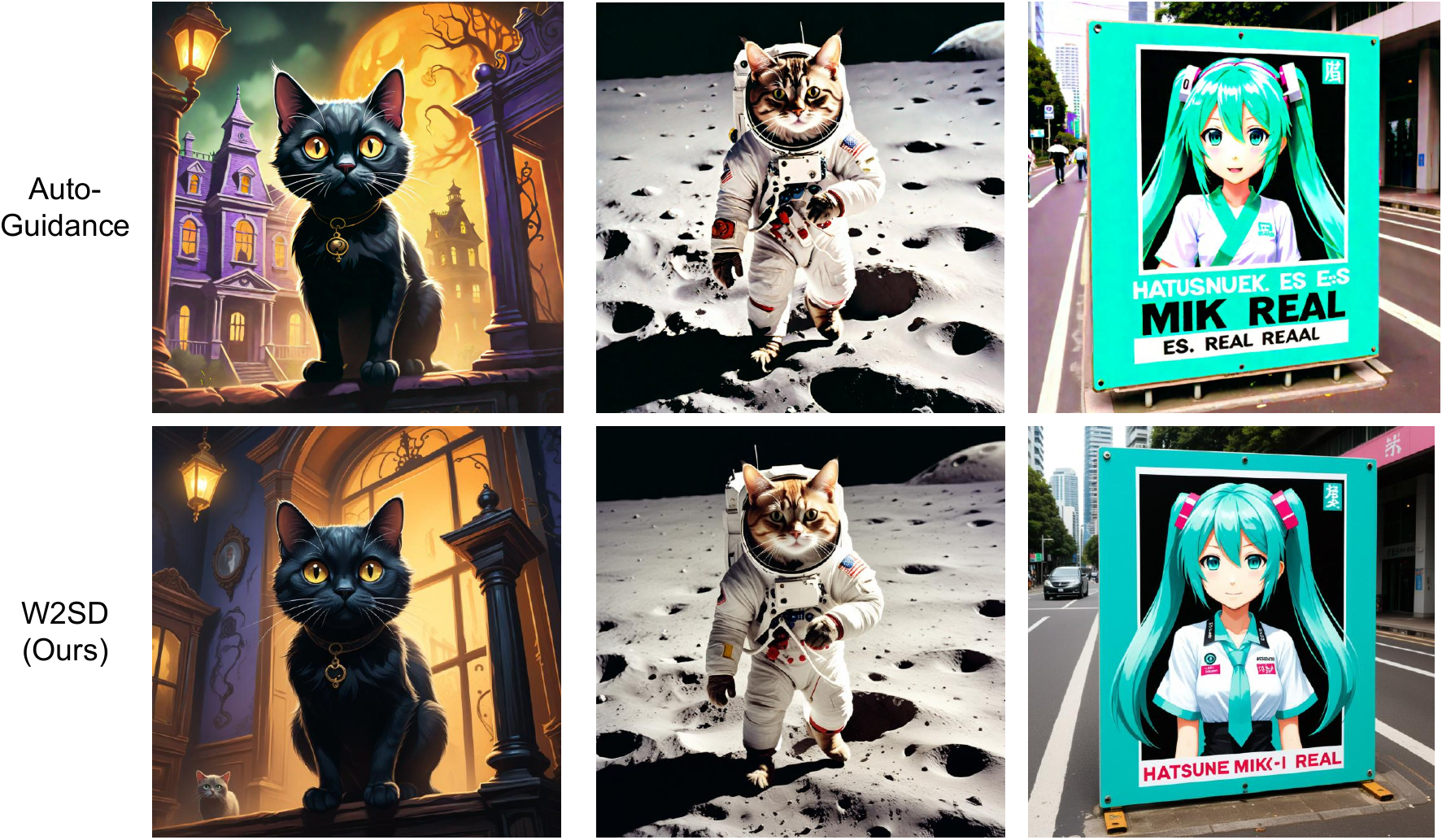}
    \caption{Compared with Auto-guidance, W2SD achieves superior performance with enhanced robustness, effectively addressing critical issues such as oversaturation and optimization failure.}
    \label{fig:augo-guidance}
\end{figure}

In W2SD's process, $x_{t} \rightarrow \textcolor{red}{x_{t-1} \rightarrow x_{t}}$, the refinement operation (marked in red) is implicitly performed by score network's internal transformation which avoids common artifacts like distortion and over-saturation, please see~\cref{fig:augo-guidance}.

On the other hand, we note W2SD generalizes the concept of weak/strong model pairs—where the "weak" model is not limited to underperforming variants created through reduced capacity or training strategies (e.g., data corruptions or degradations as in Auto-guidance).
We propose that differences in semantic interpretation of prompts or sampling methodologies (e.g., MoE routers, ControlNet adaptations) can equally constitute valid weak-strong pairings. This expanded paradigm demonstrates significantly greater practical utility, as it accommodates real-world deployment scenarios where model capabilities vary along multiple axes.

\end{document}